\documentclass{article}

\PassOptionsToPackage{numbers, compress}{natbib}
\usepackage[final,nonatbib]{neurips_2021}

\usepackage{graphicx}
\usepackage[utf8]{inputenc} 
\usepackage[numbers]{natbib}
\usepackage[T1]{fontenc}    
\usepackage[bookmarks=false,colorlinks=true,urlcolor=black,citecolor={green!60!black}]{hyperref}       
\usepackage{url}            
\usepackage{booktabs}       
\usepackage{amsfonts}       
\usepackage{nicefrac}       
\usepackage{microtype}      
\usepackage{xcolor}         
\usepackage{caption}
\usepackage{amsmath}
\usepackage{amsthm}
\usepackage{subcaption}
\usepackage{enumitem}
\usepackage{multicol}
\usepackage{multirow}
\usepackage{xspace}
\usepackage{amssymb}
\usepackage{dsfont}
\usepackage[most]{tcolorbox}
\usepackage{arydshln}

\newtheorem{theorem}{Theorem}
\newcommand{\mbr}[1]{\mathbb{R}^{#1}}
\newcommand{\semipd}[1]{\mathbb{S}_{+}^{#1}}

\DeclareMathOperator*{\argmin}{arg\,min}
\DeclareMathOperator*{\argmax}{arg\,max}
\DeclareMathOperator*{\trace}{Tr}

\newcommand{\vx}{\mathbf{x}}
\newcommand{\vy}{\mathbf{y}}
\newcommand{\mX}{\mathbf{X}}
\newcommand{\mF}{\mathbf{F}}
\newcommand{\mY}{\mathbf{Y}}
\newcommand{\mL}{\mathbf{L}}
\newcommand{\mP}{\mathbf{P}}
\newcommand{\vp}{\mathbf{p}}
\newcommand{\vv}{\mathbf{v}}
\newcommand{\vu}{\mathbf{u}}
\newcommand{\mW}{\mathbf{W}}
\newcommand{\mTheta}{\boldsymbol{\Theta}}
\newcommand{\mIdent}{\mathbf{I}}

\makeatletter
\DeclareRobustCommand\onedot{\futurelet\@let@token\bmv@onedotaux}
\def\bmv@onedotaux{\ifx\@let@token.\else.\null\fi\xspace}
\def\eg{\emph{e.g}\onedot} 
\def\ie{\emph{i.e}\onedot} 
 
\def\etc{\emph{etc}\onedot} \def\vs{\emph{vs}\onedot}
\def\wrt{w.r.t\onedot}

\makeatother

\makeatletter
\def\ps@myheadings{%
    \let\@oddfoot\@empty\let\@evenfoot\@empty
    \def\@evenhead{\thepage\hfil\slshape\leftmark}%
    \def\@oddhead{{\slshape\rightmark}\hfil\thepage}%
    \let\@mkboth\@gobbletwo
    \let\sectionmark\@gobble
    \let\subsectionmark\@gobble
    }
  \if@titlepage
  \renewcommand\maketitle{\begin{titlepage}%
  \let\footnotesize\small
  \let\footnoterule\relax
  \let \footnote \thanks
  \null\vfil
  \vskip 60\p@
  \begin{center}%
    {\LARGE \@title \par}%
    \vskip 3em%
    {\large
     \lineskip .75em%
      \begin{tabular}[t]{c}%
        \@author
      \end{tabular}\par}%
      \vskip 1.5em%
    {\large \@date \par}
  \end{center}\par
  \@thanks\@notice
  \vfil\null
  \end{titlepage}%
  \setcounter{footnote}{0}%
}
\else
\renewcommand\maketitle{\par
  \begingroup
    \renewcommand\thefootnote{\@fnsymbol\c@footnote}%
    \def\@makefnmark{\rlap{\@textsuperscript{\normalfont\color{black}\@thefnmark}}}%
    \long\def\@makefntext##1{\parindent 1em\noindent
            \hb@xt@1.8em{%
                \hss\@textsuperscript{\normalfont\@thefnmark}}##1}%
    \if@twocolumn
      \ifnum \col@number=\@ne
        \@maketitle
      \else
        \twocolumn[\@maketitle]%
      \fi
    \else
      \newpage
      \global\@topnum\z@   
      \@maketitle
    \fi
    \thispagestyle{plain}\@thanks\@notice
  \endgroup
  \setcounter{footnote}{0}%
}
\makeatother

\newcommand{\KS}[1]{[\textcolor{red}{#1}]}
\renewcommand{\comment}[1]{{}}
\newcommand{\CO}{\color{black!40!blue}}

\title{Contrastive  Laplacian Eigenmaps}

\author{
Hao Zhu$^{\dagger, \S}\quad$ Ke Sun$^{\S, \dagger}\quad$ Piotr Koniusz$\,$\thanks{The corresponding author.  $\qquad$Code:   \url{https://github.com/allenhaozhu/COLES}.\vspace{-0.5cm}}$^{\;\,,\S, \dagger}$\\
$^{\S}$Data61/CSIRO $\;\;^{\dagger}$Australian National University\\
allenhaozhu@gmail.com, sunk@ieee.org, piotr.koniusz@data61.csiro.au
}

\begin{document}

\maketitle

\begin{abstract}
Graph contrastive learning attracts/disperses node representations for similar/dissimilar node pairs under some notion of similarity. It may be combined with a low-dimensional embedding of nodes to preserve intrinsic and structural properties of a graph. In this paper, we extend the celebrated Laplacian Eigenmaps with contrastive learning, and call them COntrastive Laplacian EigenmapS (COLES). Starting from a GAN-inspired contrastive formulation, we show that the Jensen-Shannon divergence underlying many contrastive graph embedding models fails under disjoint positive and negative distributions, which may naturally emerge during sampling in the contrastive setting. In contrast, we demonstrate analytically that COLES essentially minimizes a surrogate of Wasserstein distance, which is known to cope well under disjoint distributions. Moreover, we show that the loss of COLES belongs to the family of so-called block-contrastive losses, previously shown to be superior compared to pair-wise losses typically used by contrastive methods. We show on popular benchmarks/backbones that COLES offers favourable accuracy/scalability compared to DeepWalk, GCN, Graph2Gauss, DGI and GRACE baselines.
\end{abstract}

\section{Introduction}
%

%
Celebrated graph embedding methods, including Laplacian Eigenmaps~\cite{belkin2003laplacian} and IsoMap~\cite{tenenbaum2000global}, reduce the dimensionality of the data by assuming that it lies on a low-dimensional manifold. 
The objective functions used  in studies \cite{belkin2003laplacian,tenenbaum2000global} model the pairwise node similarity~\cite{cai2018comprehensive} by encouraging the embeddings of nodes to lie close in the embedding space if the nodes are closely related. 
In other words, such penalties do not guarantee that unrelated graph nodes are separated from each other in the embedding space. 
%
%
%
For instance, Elastic Embedding \cite{elastic_net} uses data-driven  affinities for the so-called local distance term and the data-independent repulsion term. 

In contrast, modern graph embedding models, often unified under the Sampled Noise Contrastive Estimation (SampledNCE) framework~\cite{mikolov2013distributed,levy2014neural} and extended to graph  learning \cite{tang2015line,hamilton2017inductive,yang2020understanding}, enjoy contrastive objectives. 
%
%
%
%
By maximizing the mutual information between  patch representations and  high-level summaries of the graph, Deep Graph Infomax (DGI)~\cite{velivckovic2017graph} is a contrastive method. GraphSAGE \cite{hamilton2017inductive} 
minimizes/maximizes distances between so-called positive/negative pairs, respectively. It relies on the inner product passed through the sigmoid non-linearity, which we argue below as suboptimal. 
%
%
%

\definecolor{beaublue}{rgb}{0.9, 0.95, 0.9}
\definecolor{blackish}{rgb}{0.2, 0.2, 0.2}
\vspace{-0.05cm}
\begin{tcolorbox}[width=1.0\linewidth, colframe=blackish, colback=beaublue, boxsep=0mm, arc=2mm, left=2mm, right=2mm, top=5mm, bottom=2mm]
\vspace{-0.3cm}
Thus, we propose a new \textbf{COntrastive Laplacian EigenmapS (COLES)} framework for unsupervised network embedding. 
COLES, derived from SampledNCE framework  \cite{mikolov2013distributed,levy2014neural}, realizes the negative sampling strategy for Laplacian Eigenmaps. 
%
%
Our general objective is given as:
\vspace{-0.1cm}
\begin{equation}
\mTheta^* = \argmax\limits_{\mTheta} \trace(f_{\mTheta}(\mX)^\top\Delta\mathbf{W}f_{\mTheta}(\mX))+\beta\Omega(f_{\mTheta}(\mX)).
\label{eq:COLESGNN}
\vspace{-0.05cm}
\end{equation}
\end{tcolorbox}
\vspace{-0.1cm}
$\mX\in\mbr{n\times d}$ in Eq. \eqref{eq:COLESGNN} is the node feature matrix with $d$ feature dimensions given $n$ nodes, $f_{\mTheta}(\mX)\in \mbr{n\times d'}$ is an output of a chosen Graph Neural Network backbone (embeddings to optimize) with the feature dimension $d'$, $\mTheta$ denotes network parameters, whereas $\Delta\mathbf{W}\in\semipd{n}$ is the difference between the degree-normalized positive and negative adjacency matrices which represent the data graph and some negative graph capturing negative links for contrastive learning.
\begin{figure}[t]
 \centering
     \begin{subfigure}[b]{0.3\textwidth}
         \centering
         \includegraphics[width=4.5cm]{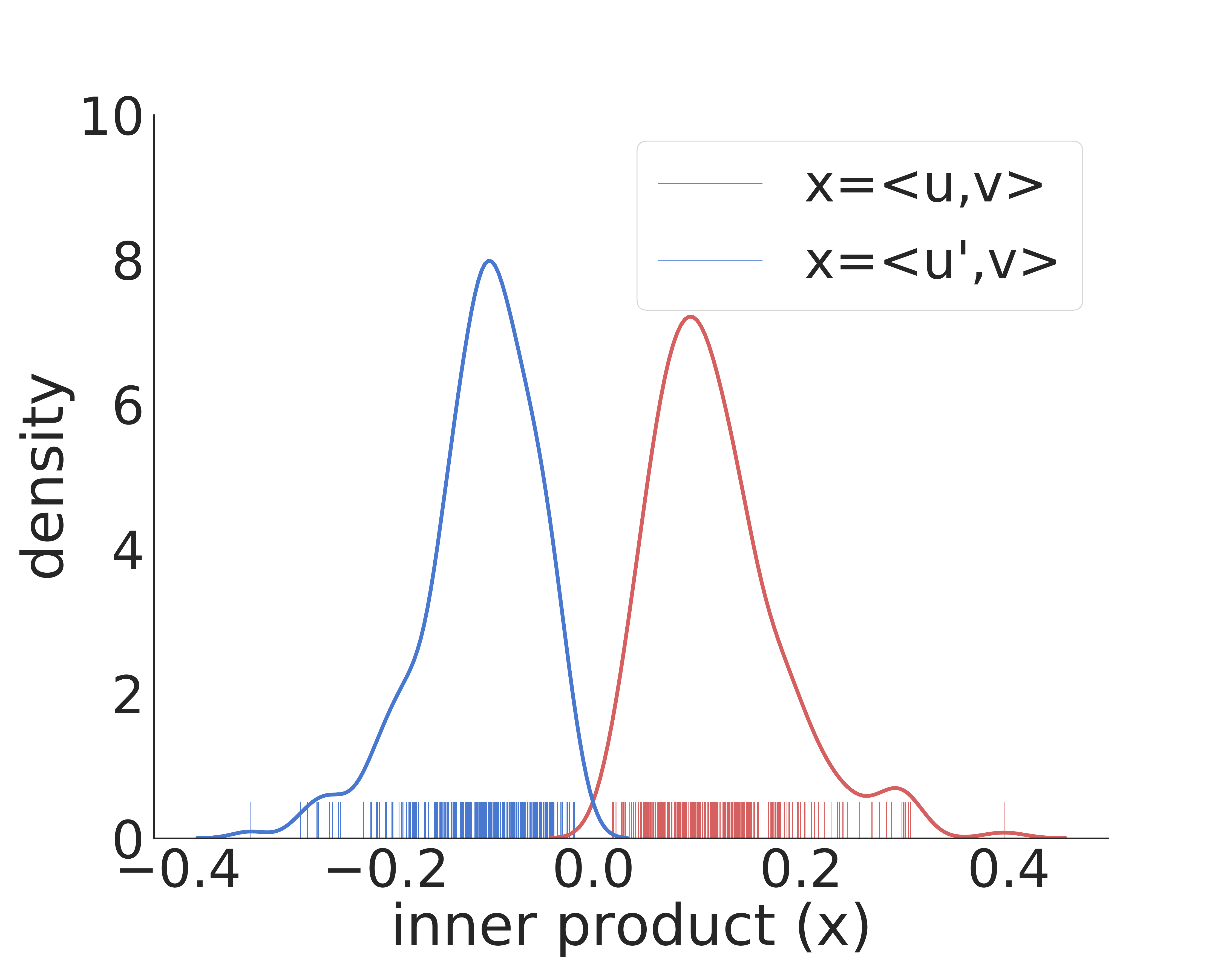}
         \caption{}
         \label{fig:optD}
     \end{subfigure}
     \begin{subfigure}[b]{0.3\textwidth}
         \centering
         \includegraphics[width=4.5cm]{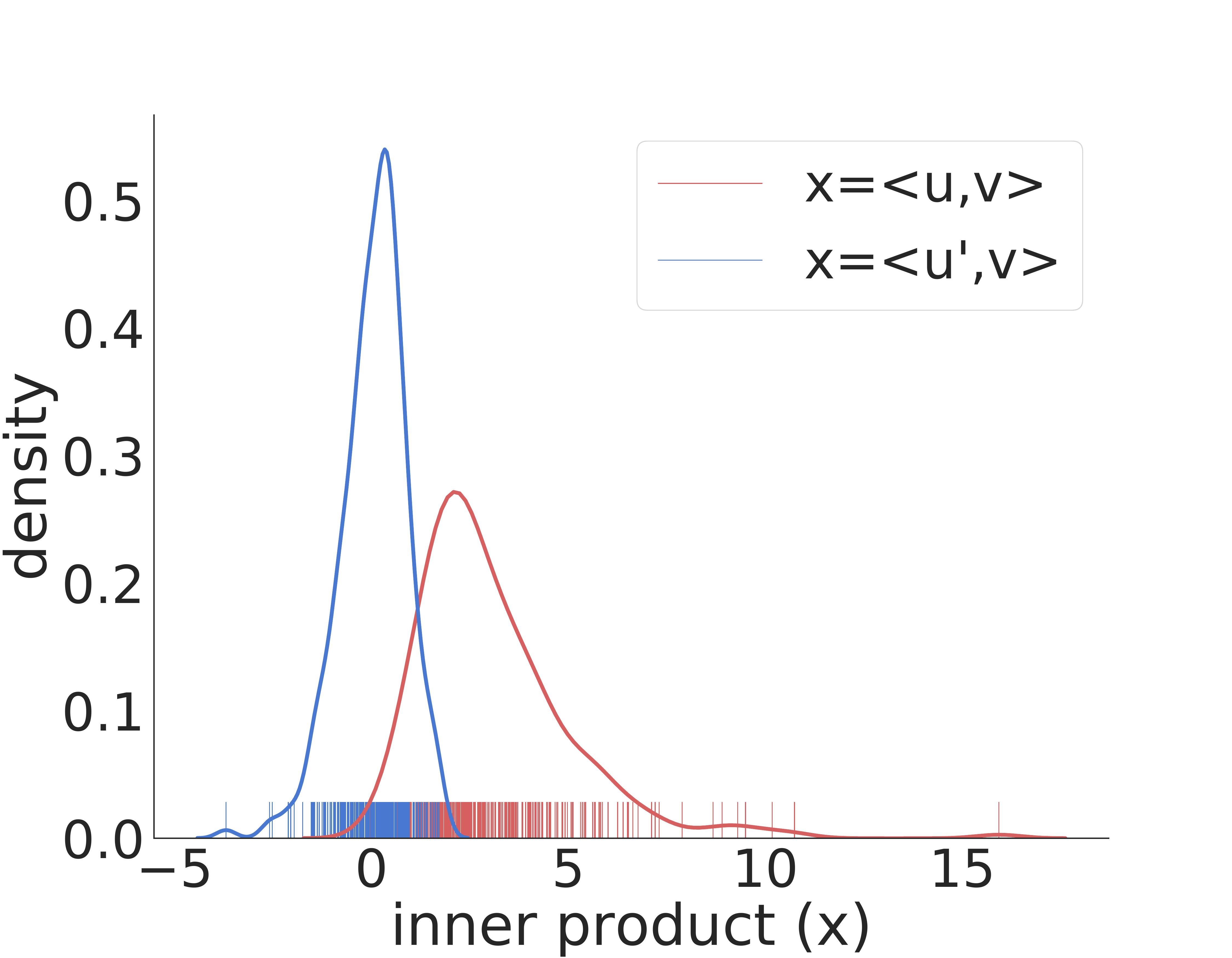}
         \caption{}
         \label{fig:nonoptD}
     \end{subfigure}
\caption{Densities of dot-product scores $\langle\vv,\vu\rangle$ and $\langle\vv,\vu'\rangle$  (red and blue curves) between the anchor/positive embedding and the anchor/negative embedding (GCN contrastive setting). Left/right figures use two distinct minibatches sampled on Cora. With the small overlap of distributions, many contrastive methods relying on the JS divergence may underperform (see Section \ref{sec:wass} for details).}
\label{fig:mot}
\end{figure}
Moreover, $\beta\geq0$ controls the regularization term $\Omega(\cdot)$ whose role is to constrain the $\ell_2$ norm of network outputs or encourage the so-called incoherence \cite{incoherence} between column vectors. Section \ref{sec:col_lin} presents COLES for the Linear Graph Network (LGN) family, in which we take special interest due to their simplicity and agility.



%
By building upon previous studies \cite{levy2014neural,arjovsky2017wasserstein,weng2019gan}, we show that COLES can be derived by reformulating SampledNCE into Wasserstein GAN using a GAN-inspired contrastive formulation. This result has a profound impact on the performance of COLES, as 
the standard contrastive approaches based on SampledNCE strategy (\ie,  GraphSAGE \cite{hamilton2017inductive}) turn out to utilize the Jensen-Shannon divergence, which yields $\log 2$ constant and vanishing gradients for disjoint distributions of positive and negative sampled pairs used for contrastive learning. Figure \ref{fig:mot} shows two examples of such nearly disjoint distributions. In contrast, COLES by design  avoids the sigmoid in favour of the Radial Basis Function (RBF) non-linearity. We show that such a choice coincides with  a surrogate of Wasserstein distance, 
which is known for its robustness under poor overlap of distributions, leading to the good performance of COLES. Moreover, we also show that the loss of COLES belongs to the family of so-called block-contrastive losses, which were shown to be superior compared to pair-wise losses \cite{saunshi2019theoretical}.
%
%
%
%
%
%
In summary, our contributions are threefold:
\renewcommand{\labelenumi}{\roman{enumi}.}
\vspace{-0.05cm}
\hspace{-1.0cm}
\begin{enumerate}[leftmargin=0.6cm]
\item We derive COLES, a reformulation of the Laplacian Eigenmaps into a contrastive setting, based on the SampledNCE framework \cite{mikolov2013distributed,levy2014neural}.
\item By using a formulation inspired by GAN, we show that COLES essentially minimizes a surrogate of Wasserstein distance, as opposed to the Jensen-Shannon (JS) divergence emerging in traditional contrastive learning. Specifically, by showing the Lipschitz continuous nature of our formulation, we prove that 
our formulation enjoys the Kantorovich-Rubinstein duality for the Wasserstein distance.
\item We show COLES enjoys a block-contrastive loss known to outperform pair-wise  losses \cite{saunshi2019theoretical}.
\end{enumerate}
\noindent\textbf{Novelty.} 
We propose a simple way to obtain contrastive parametric graph embeddings which works with numerous backbones. For instance, we  obtain spectral graph embeddings by combining COLES with SGC \cite{wu2019simplifying} and S\textsuperscript{2}GC~\cite{zhu2021simple}, which is solved by the SVD decomposition.

%
%
%
%

\section{Preliminaries}
\paragraph{Notations.} Let $G\!=\!(V, E)$ be a simple, connected and undirected graph with $n\!=\!|V|$ nodes and $m\!=\!|E|$ edges.
%
%
Let $i\in\{1,\cdots, n\}$ be the node index of $G$, and $d_j$
be the degree of node $j$ of $G$.
Let $\mathbf{W}$ be the adjacency matrix, and $\mathbf{D}$ be the diagonal matrix containing degrees of nodes. 
%
%
Moreover, let $\mathbf{X} \in \mbr{n\times d}$ denote the node feature matrix where each node $v$ is associated with a feature vector $\mathbf{x}_v\in \mbr{d}$. 
Let the normalized graph Laplacian matrix be defined as $\mL = \mathbf{I}-\mathbf{D}^{-1/2}\mathbf{\widehat{W}}\mathbf{D}^{-1/2}\in\semipd{n}$,  a symmetric positive semi-definite matrix. 
Finally, scalars and  vectors  are denoted by lowercase regular and bold fonts, respectively. Matrices are denoted by uppercase bold fonts. 
%

\subsection{Negative Sampling}

SampledNCE \cite{gutmann2010noise,mikolov2013distributed,levy2014neural}, a contrastive learning framework, is used by numerous works \cite{tang2015line,hamilton2017inductive,yang2020understanding}. Let
%
$p_d(u|v)$ and $p_n(u'|v)$ be the so-called data and negative distributions given the so-called anchor node $v$, where $u$ and $u'$ denote the node for a positive and negative sample, respectively. Let $p_{d}(v)$ be the anchor distribution. Given some loss components $s_{\mTheta}(v, u)$ and  $\tilde{s}_{\mTheta}(v, u')$ whose role is to evaluate the similarity for pairs $(v,u)$ and $(v,u')$, the contrastive loss is typically given as:
\begin{equation}
J(\mTheta)=\mathbb{E}_{v \sim p_{d}(v)}\left[\mathbb{E}_{u \sim p_{d}(u \mid v)} s_{\mTheta}(v, u) + \eta\mathbb{E}_{u^{\prime} \sim p_{n}\left(u^{\prime} \mid v\right)} \tilde{s}_{\mTheta}(v, u')\right],
\label{eq:seploss}
\end{equation}
where $\eta>0$ controls the impact of negative sampling. 
Let $\vu\in\mbr{d'}$ be the embedding of the node $u$ obtained with an encoder $f_{\mTheta}(\vx_u)$ given parameters $\mTheta$, where $\vx_u\in\mbr{d}$ is the initial node feature vector. 
Let $\vu'\in\mbr{d'}$ and $\vv\in\mbr{d'}$ be embeddings of nodes $u'$ and $v$, accordingly. Let $s_{\mTheta}(u,v) =\log\sigma(\mathbf{u}^{\top} \mathbf{v})$ and $\tilde{s}_{\mTheta}(u^{\prime},v)=\log (1\!-\!\sigma(\mathbf{u}'^\top\mathbf{v}))$, where $\sigma(\cdot)$ is the sigmoid function. 
Subsequently, one obtains the contrastive objective (to be maximized), employed by LINE \cite{tang2015line}, REFINE~\cite{zhu2021refine}, GraphSAGE \cite{hamilton2017inductive} and many other methods according to  \citet{yang2020understanding}:
\begin{equation}
J(\mTheta) =\mathbb{E}_{v \sim p_{d}(v)}\left[\mathbb{E}_{u \sim p_{d}(u \mid v)} \log \sigma(\mathbf{u}^{\top} \mathbf{v})+\eta\mathbb{E}_{u' \sim p_{n}\left(u' \mid v\right)} \log \sigma(-\mathbf{u}'^\top\mathbf{v})\right].
\label{eq:nice}
\end{equation}
In what follows, we argue that the choice of sigmoid for $\sigma(\cdot)$ leads to negative consequences. Thus, we derive COLES under a different choice of $s_{\mTheta}(v, u)$ and $\tilde{s}_{\mTheta}(v, u')$.

\section{Methodology}

In what follows, we depart from the above setting of (typical) contrastive sampling, which results in a derivation of our COntrastive Laplacian EigenmapS (COLES).

\subsection{Contrastive Laplacian Eigenmaps}

\comment{
\KS{
Assume:
\begin{equation}
\forall{u},
\Vert{u}\Vert=1,
\quad
E_p(\mathrm{trace}(({L}-\eta'{L}^{-})u u^T)) \ge -\epsilon
\end{equation}
where $\mL$ and $\mL^{-}$ are random matrices (corresponding to the sampling process), $\epsilon>0$, $E_p$ is the expectation.
This is an additional constraint of the sampling process, besides the current $Y^\top{}Y=I$.
}
\KS{
Consider the example $L$ is the symmetic graph Laplacian,
The spectrum of $\rho(L)\subset[0,2]$.
$L^{-}=I-\frac{1}{N}11^\top$ corresponds to the simplest negative sampling (uniform distribution).
The spectrum of $L^{-1}$ is simply $\{1/N\}$.
Therefore we have
\begin{equation}
E_p(\mathrm{trace}(({L}-\eta'{L}^{-})u u^T)) \ge -\frac{\eta'}{N}
\end{equation}
Need another example of $L^{-}$ in closed form.
}
}

%
%
%
\comment{
\begin{equation}
\!\!\!\!\!\!\!\!J(\mTheta)
=\mathbb{E}_{v \sim p_{d}(v)}[\mathbb{E}_{u \sim p_{d}(u \mid v)} \left\|\mathbf{u}-\mathbf{v}\right\|^2_2 -k \mathbb{E}_{u^{\prime} \sim p_{n}\left(u^{\prime} \mid v\right)} \left\|\mathbf{u}^{\prime} - \mathbf{v}\right\|^2_2],
\label{eq:dist}
\end{equation}
}
Instead of log-sigmoid used in $s_{\mTheta}(v, u)$ and $\tilde{s}_{\mTheta}(v, u')$ of Eq. \eqref{eq:nice}, let us substitute $s_{\mTheta}(v, u) = \log\exp(\mathbf{u}^{\top} \mathbf{v})=\mathbf{u}^{\top} \mathbf{v}$  and $\tilde{s}_{\mTheta}(v, u')=\log\exp(-\mathbf{u}'^\top\mathbf{v})=-\mathbf{u}'^\top\mathbf{v}$ into Eq.~\eqref{eq:seploss}, which yields:
\begin{equation}
\begin{aligned}
\!\!J(\mTheta)\!=\!\mathbb{E}_{v \sim p_{d}(v)}\left[\mathbb{E}_{u \sim p_{d}(u \mid v)} (\mathbf{u}^{\top} \mathbf{v})+\eta\mathbb{E}_{u'\sim p_{n}\left(u'\mid v\right)} \left(-\mathbf{u}'^\top\mathbf{v}\right)\right].
\label{eq:final1}
\end{aligned}
\end{equation}
We assume that variables of the above  objective (to maximize) can be constrained (\eg, by the $\ell_2$ norms to prevent ill-posed solutions) and represented by degree-normalized adjacency matrices. Next, we cast Eq. \eqref{eq:final1} into the objective of COLES (refer to our Suppl. Material for derivations):
\begin{equation}
\begin{aligned}
   \mathbf{Y}^{*} &=\argmin\limits_{\mY,\text{ s.t. }\mY^\top\!\mY=\mIdent} \trace(\mathbf{Y}^\top\mL\mathbf{Y}) - \frac{\eta'}{\kappa}\sum_{k=1}^\kappa \trace(\mathbf{Y}^\top{\mL_k^{(-)}}\mathbf{Y})\\
   &=\argmax\limits_{\mY,\text{ s.t. }\mY^\top\!\mY=\mIdent} \trace(\mathbf{Y}^\top\!\Delta\mathbf{W}\mathbf{Y}) \quad\text{ where }\quad \Delta\mathbf{W}\!=\!\mathbf{W}^{(+)}-\frac{\eta'}{\kappa}\sum\limits_{k=1}^\kappa\!\mathbf{W}_k^{(-)},
\end{aligned}
\label{eq:CL}
\end{equation}
and the rows of matrix $\mY\in\mbr{n\times d'}$ contain the embedding vectors,  $\mL_k^{(-)}$ for $k=1,\cdots,\kappa$ are randomly generated degree-normalized  Laplacian matrices capturing the negative sampling,  $\mL_k^{(-)}\!=\!\mathbf{I}\!-\!\mathbf{W}_k^{(-)}$ and $\mathbf{L}\!=\!\mathbf{I}\!-\!\mathbf{W}^{(+)}$.
The scalar $0\leq\eta'\leq 1$  
ensures that 
$\mL\!-\!\frac{\eta'}{\kappa}\sum_{k=1}^{\eta'}\mL_k^{(-)}\in\semipd{n}$ (one could truncate the negative spectrum instead) and controls the impact of $\mathbf{W}_k^{(-)}$.  

We note that COLES minimizes over the standard Laplacian Eigenmap while maximizing over the randomized Laplacian Eigenmap, which alleviates the lack of negative sampling in the original  Laplacian Eigenmaps. 
However, unlike Laplacian Eigenmaps, we do not optimize over free variables $\mY$ but over the network parameters, as in Eq. \eqref{eq:COLESGNN} and \eqref{eq:COLES_linear}.
%
%
%
Clearly, if $\eta'\!=\!0$ and $\mathbf{Y}$ are free variables, Eq. \eqref{eq:CL} reduces to standard Laplacian Eigenmaps \cite{belkin2003laplacian}: $\mathbf{Y}^{*}\!=\argmin_{\mY, \text{ s.t. }\mY^\top\mY=\mIdent} \trace(\mathbf{Y}^\top\mL\mathbf{Y})$. 
%
%

\vspace{-0.0cm}
\begin{tcolorbox}[width=1.0\linewidth, colframe=blackish, colback=beaublue, boxsep=0mm, arc=2mm, left=2mm, right=2mm, top=5mm, bottom=2mm]
\vspace{-0.2cm}
\paragraph{COLES for Linear Graph Networks.}
\label{sec:col_lin}
$\!\!\!\!$In what follows, we are especially interested in the lightweight family of LGNs such as SGC \cite{wu2019simplifying} and S\textsuperscript{2}GC~\cite{zhu2021simple} (APPNP~\citep{klicpera2018predict} with the linear activation could be another choice) whose COLES-based objective can be reformulated as:$\!\!\!$
\vspace{-0.0cm}
\begin{equation}
\mP^* = \argmax\limits_{\mP, \text{ s.t. }\mP\mP^\top=\mIdent} \trace(\mP\mX^\top\mF^\top\!\Delta\mathbf{W}\mF\mX\mP^\top).
\label{eq:COLES_linear}
\vspace{-0.0cm}
\end{equation}
\end{tcolorbox}

\noindent$\mF\in\mbr{n\times n}$ in Eq. \eqref{eq:COLES_linear} is the so-called spectral filter operating on the (degree-normalized) graph adjacency matrix, and  $\mP\in\mbr{d'\times d}$ is a unitary projection matrix such that $0<d'<d$. The solution to Eq. \eqref{eq:COLES_linear} can be readily obtained by solving the generalized eigenvalue problem $\mX^\top\mF^\top\!\Delta\mathbf{W}\mF\mX\vp=\lambda\vp$ (an SVD on a small $d\times d$ matrix $(\mX^\top\mF^\top\!\Delta\mathbf{W}\mF\mX)\in\semipd{d}$). This step results in a matrix of embeddings $f_\mP(\mX)=\mF\mX\mP^\top\in\mbr{n\times d'}$ for supervised training. Based on given a degree-normalized graph adjacency matrix $\mW\in\mbr{n\times n}$, the spectral filters 
for SGC and S\textsuperscript{2}GC are given as $\mW^{K'}$ and $\alpha\mIdent+\frac{1-\alpha}{K'}\sum_{k=1}^{K'}\mW^k$. Here, integer $K'\geq 1$ and scalar $\alpha\geq 0$ are the number of layers and the importance of self-loop. Note that Eq. \eqref{eq:COLES_linear} is related to Locality Preserving Projections \cite{he2004locality} if $\eta'=0$. Note also that enforcing the orthogonality constraints in Eq. \eqref{eq:COLES_linear} coincides with the SVD-based solution described above. In contrast, the more general form of COLES in Eq. \eqref{eq:COLESGNN} requires the regularization or constraints (depending on the backbone) imposed on minibatches $i\in\mathcal{B}$  \eg, we used the soft penalty $\Omega(f_{\mTheta}(\mX_{i}))\!=\!\lVert f^\top_{\mTheta}(\mX_{i})f_{\mTheta}(\mX_{i})-\mIdent\rVert_F^2$. 

\vspace{-0.0cm}
\begin{tcolorbox}[width=1.0\linewidth, colframe=blackish, colback=beaublue, boxsep=0mm, arc=2mm, left=2mm, right=2mm, top=5mm, bottom=2mm]
\vspace{-0.2cm}
\vspace{-0.2cm}
\paragraph{COLES (Stiefel).} 
Inspired by the Locality Preserving Projections \cite{he2004locality} and Eq. \eqref{eq:COLES_linear}, we also investigate:  
\vspace{-0.2cm}
\begin{equation}
(\mP^*,\mTheta^*) = \argmax\limits_{\mP,\mTheta, \text{ s.t. }\mP\mP^\top=\mIdent} \trace(\mP f^\top_{\mTheta}(\mX)\Delta\mathbf{W}f_{\mTheta}(\mX)\mP^\top),
\label{eq:COLES_stiefel}
\vspace{-0.0cm}
\end{equation}
solved on the Stiefel manifold by GeoTorch \cite{NEURIPS2019_stiefel}. The embed. is: $f_\mP(\mX)=f_{\mTheta}(\mX)\mP^\top\in\mbr{n\times d'}$. 
\end{tcolorbox}

\comment{
\subsection{Laplacian Eigenmaps}
Graph embedding  yields a low-dimensional embedding of nodes which preserves the connectivity between pairs of nodes on the graph. 
We denote the low-dimensional embedding of the nodes as $\mathbf{Y}=[\vy_1;\cdots;\vy_n]$, where the row vector $\vy_i$ is the embedding of node $i$. Direct graph embedding aims to maintain similarities among node pairs according to the graph preserving criterion:
\begin{align}
    \!\!\!\!\!\mY^*\!=\!\argmin\limits_{\vy_1,\cdots,\vy_n} \sum_{(i,j)\in E}\|\mathbf{y}_i-\mathbf{y}_j\|_2^2\,W_{ij} 
    \!=\!\argmin\limits_{\mY} \trace(\mY^\top\mathbf{L}\mY),\!\!
    \label{eq:lap}
\end{align}
where 
$W_{ij}$ is the (weighted) adjacency matrix defining connectivity between two nodes according to similarity between pairs of feature vectors $\vx_i$ and $\vx_j$. Moreover,  minimizing over $\mY$ is typically s.t.  $\trace(\mathbf{Y}^\top\mathbf{D}\mathbf{Y})\!=\!1$. 
The graph preserving criterion of Laplacian Eigenmaps assumes that as the distance between  feature vectors $\mathbf{x}_i$ and $\mathbf{x}_j$ is smaller, then the distance between embeddings $\mathbf{y}_i$ and $\mathbf{y}_j$ will be accordingly smaller as well. However, if the similarity between two nodes is zero, embeddings $\mathbf{y}_i$ and $\mathbf{y}_j$ are not explicitly minimized to capture such cases. 
%
%
The above approach requires that all the nodes are available at the time of solving Eq. \eqref{eq:lap}. 
Locality Preserving Projections (LPP) of \cite{he2004locality} provide an extension which results in a projection matrix to let one embed feature vectors of nodes  unseen during the minimization step.
}

\comment{
\paragraph{Locality Preserving Projections (LPP).} This approach assumes that a low-dimensional vector representation can be obtained by a linear projection  $\mY\!=\!\mX\mP^{*\top}$. Instead of the objective function Eq. \eqref{eq:lap}, the authors solve:
\begin{equation}
\begin{aligned}
   \mathbf{P}^{*}\!\!=\!\argmin\limits_{\mathbf{P}}\!\!\!\sum_{(i,j)\in E}\!\!\!\|\mathbf{P}^\top(\vx_i-\vx_j)\|^2_2\,W_{ij}\!=\!\argmin\limits_{\mathbf{P}} \trace(\mP\mX^\top\mL\mX\mP^\top).
\end{aligned}
\label{eq:lpp}
\end{equation}
Commonly, the above linearized extension of Laplacian Embeddings is computationally efficient for training and testing stages. However, its performance may degrade on cases with non-linearly separable data. Thus,  we introduce a GNN-based extension  to handle non-linear data. 
}

\section{Theoretical Analysis}
\subsection{COLES is Wasserstein-based Contrastive Learning} \label{sec:wass}
\newcommand{\bm}{\mathbf}


By casting the positive and negative distributions of SampledNCE as the real and generated data distributions of GAN, the key idea of this analysis is to (i) cast the traditional contrastive loss in Eq. \eqref{eq:nice} (used by LINE~\cite{tang2015line}, GraphSAGE \cite{hamilton2017inductive} and other methods \cite{yang2020understanding}) as a GAN framework, and show this corresponds to the use of JS divergence and (ii) cast the objective of COLES in  Eq. \eqref{eq:final1} as a GAN framework, and show it corresponds to the use of
 a surrogate of Wasserstein distance. The latter outcome is preferable under the vanishing overlap of two distributions, as the JS divergence  yields $\log(2)$ constant and vanishing gradients. The Wasserstein distance suffer less from this issue.

%
%
%
%

 
For simplicity, consider the embedding $\bm{v}$ of the anchor node is given.
An embedding vector $\mathbf{u}$ is sampled from the `real' distribution
$p_r(\mathbf{u}) = p_{d}(u \mid v)$,
and $\mathbf{u}'$ is sampled from the `generator' distribution
$p_g(\mathbf{u}) = p_{n}\left(u' \mid v\right)$.
Following \citet{arjovsky2017wasserstein} and \citet{weng2019gan}, one arrives at a GAN-inspired formulation
which depends on the choice of `discriminator' $D(\mathbf{u})$:
\begin{equation}
    \max_{\bm\Theta}
    \int_{\mathbf{u}} \bigg( p_r(\mathbf{u}) \log(D(\mathbf{u}))
                               + p_g(\mathbf{u}) \log(1 - D(\mathbf{u}))
                       \bigg) \mathrm{d}\mathbf{u}
\le
     2\mathrm{JS} \left(p_{r} \| p_{g}\right) -2 \log 2,
\label{eq:ganlike}
\end{equation}
where $\mathrm{JS}\left(p_{r} \| p_{g}\right)$ denotes the
Jensen-Shannon (JS) divergence.
If $D(\bm{u})$ is completely free, then the optimal
$D^*(\bm{u})$ which maximizes the left-hand-side (LHS) of Eq.~\eqref{eq:ganlike} is $D(\bm{u}) = p_r(\bm{u})/(p_r(\bm{u})+p_g(\bm{u}))$.
%
%
Plugging $D^*$ back into the LHS,
we get the right-hand-side (RHS) of the inequality.
In our setting, the case $p_g\sim p_r$ means that negative sampling yields hard negatives, that is, negative and positive samples are very similar.
Hence, this family of embedding techniques
try to optimally discriminate $p_r$ and $p_g$ in the embedding space.


The above analysis shows that traditional contrastive losses
are bounded by the JS divergence.
Regardless of the choice of $D(\bm{u})$,
if the support of the density $p_r$ and the support of $p_g$
are disjoint (\eg, positive and negative samples in the minibatch of the SGD optimization), the JS divergence yields zero and vanishing gradients. 
If the `discriminator' is set to $D(\mathbf{u})=\sigma(\mathbf{u}^{\top}\mathbf{v})$,
the objective in Eq.~\eqref{eq:ganlike} becomes exactly Eq.~\eqref{eq:nice}.
By noting $\partial\log\sigma(\bm{u}^\top\bm{v})/\partial\bm{u}=
\sigma(\bm{u}^\top\bm{v})\bm{v}$, the gradient is likely to
vanish due to the scalar $\sigma(\bm{u}^\top\bm{v})$
and does not contribute to learning of network parameters.
Figure \ref{fig:mot} shows densities of $x\!=\vu^\top\vv$ and $x\!=\vu'^\top\vv$ for 
$p_r$ and $p_d$ estimated by the Parzen window on two sampled minibatches of contrastive GCN. Clearly, these distributions are approximately disjoint.


Compared with the JS divergence, the Wasserstein distance
considers the metric structure of the embedding space:
\vspace{-0.2cm}
\begin{equation}
    \inf_{\gamma \sim \Pi(p_r, p_g)}
    \mathbb{E}_{(\bm{u}, \bm{u}') \sim \gamma}\| \bm{u}-\bm{u}'\|_1,
\label{eq:ws}
\end{equation}
where  $\Pi(p_r, p_g)$ is the set of joint distributions with marginals $p_r(\bm{u})$ and $p_g(\bm{u}')$.

\vspace{-0.0cm}
\begin{tcolorbox}[width=1.0\linewidth, colframe=blackish, colback=beaublue, boxsep=0mm, arc=2mm, left=2mm, right=2mm, top=5mm, bottom=2mm]
\vspace{-0.2cm}
By the Kantorovich-Rubinstein duality~\cite{ot}, 
\textbf{the optimal transport problem for COLES} can be equivalently expressed as:
\begin{equation}
\begin{aligned}
    & \sup_{g:\,K(g)\le1}
       \left(
           \mathbb{E}_{\bm{u} \sim p_r}[g(\bm{u})]
        - \mathbb{E}_{\bm{u}'\sim p_g}[g(\bm{u}')]
    \right)
    \\ &\geq
    \max_{\mTheta}
    \left[\mathbb{E}_{u \sim p_{d}(u \mid v)}
    (\mathbf{u}^{\top} \mathbf{v})+ \mathbb{E}_{u' \sim p_{n}(u' \mid v)} (-\mathbf{u}'^{\top} \mathbf{v})\right],
\end{aligned}
\label{eq:kant}
\end{equation}
under a drawn anchor $v\sim p_d(v)$,
where $K(g)$ means the Lipschitz constant,
and supreme is taken over all 1-Lipschitz functions (or equivalently, all
$K$-Lipschitz functions.) 
\end{tcolorbox}

The ``$\ge$'' is because $g(\bm{u})$ is chosen to
the specific form $g_v(\vu)=\vu^\top\vv$,
where $\vv$ is parameterized by a graph neural network with parameters
$\mTheta$.
Optimizing over the neural network parameters $\bm\Theta$
can enumerate a subset of functions which satisfies the Lipschitz constant $K$.

%
%


\noindent\textbf{Lipschitz continuity of COLES.} 
In order to assure the Lipschitz continuity of COLES,
let individual embeddings be stacked row-wise into a matrix and $\ell_2$-norm normalized along rows, or 
along columns.
Given $v$ (the reference node), the following holds:
\begin{equation*}
\vert \vu^\top\vv-\vu'^\top\vv\vert
\le
\Vert \vv \Vert_{\max}
\Vert\vu - \vu' \Vert_1,
\end{equation*}
where
$K = \max_v \Vert\vv\Vert_{\max}$ ($\le1$ in the case of either sphere embedding or the constraint $\mY^\top\mY=\mIdent$ of the COLES formula in Eq. \eqref{eq:CL}).
Thus, the function
$g(\vu)=\vu^\top\vv$ is Lipschitz with constant $K$. 

%


\subsection{COLES enjoys the Block-contrastive Loss}
\label{sec:contr}


We notice that COLES leverages an access to blocks of similar data, rather than just individual pairs in the loss function. 
To this end, we resort to the Prop. 6.2 of 
\citet{saunshi2019theoretical}, which shows that for family of functions $\mathcal{F}$ whose $\lVert f(\cdot)\rVert\leq R$ for some $R>0$,  a block-contrastive loss $L_{u n}^{\text {block }}$ is always bounded by a pairwise-contrastive loss $L_{\text {un }}$, that is, $L_{u n}^{\text {block }}(f)\leq L_{\text {un }}(f)$. To that end, 
\citet{saunshi2019theoretical} also show that   as block-contrastive losses achieve lower minima than their pairwise-contrastive counterparts, they also enjoy better generalization. 

\comment
{
%
%
\begin{theorem}
$\forall f \in \mathcal{F}$, we have:
\begin{equation}
\!\!\!\!\!\!\!\!\!\!\!L_{sup}(f)\leq
\frac{1}{1-\tau}\left(L_{u n}^{\text {block }}(f)-\tau\right) \leq \frac{1}{1-\tau}\left(L_{\text {un }}(f)-\tau\right),
\end{equation}
where $\tau\!=\!\!\!\!\!\!\!\underset{c, c' \sim \rho^{2}}{\mathbb{E}} \!\!\!\delta(c\!-\!c')$ be the probability that two classes
sampled independently from $\rho$ are the same, $\delta(c\!-\!c')\!=\!1$ if $c$ and $c'$ are the same, 0 otherwise. Distribution $\rho$ over the classes that characterizes how these classes naturally occur in the unlabeled data. 
\label{th:bc}
\end{theorem}
}

We show that COLES is a block-contrastive loss, which explains its good performance.  Following Eq. \eqref{eq:final1}, for a given embedding $\mathbf{v} = f_{\mTheta}(\mathbf{x}_v)$, and $b$ embeddings $\mathbf{u}_i = f_{\mTheta}(\mathbf{x}_{u_i})$ and $\mathbf{u}'_i = f_{\mTheta}(\mathbf{x}_{u'_{i}})$ drawn according to $p_{d}(u \mid v)$ and $p_{n}(u' \mid v)$, we have (note minus preceding eq. as here we minimize):
\vspace{-0.1cm}
\begin{equation}
\!\!\!\!-\mathbb{E}_{u \sim p_{d}(u \mid v)} (\mathbf{u}^{\top} \mathbf{v})+\mathbb{E}_{u'\sim p_{n}\left(u'\mid v\right)} \left(-\mathbf{u}'^\top\mathbf{v}\right)=
-\mathbf{v}^\top\left(\frac{\sum_{i}\mathbf{u}_i}{b}-\frac{\sum_{i} \mathbf{u}^\prime_i}{b'}\right)=-\mathbf{v}^\top(\boldsymbol{\mu}^{+}-\boldsymbol{\mu}^{-}),
\label{eq:block}
\vspace{-0.1cm}
\end{equation}
where $\boldsymbol{\mu}^{+}$ and $\boldsymbol{\mu}^{-}$ are positive and negative block summaries of sampled nodes. Looking at Eq. \eqref{eq:CL}, it is straightforward to simply expand $\sum_{(i,j)\in E}\|\mathbf{y}_i-\mathbf{y}_j\|_2^2\,\Delta W_{ij}$ to see that each index $i$ will act as a selector of anchors, whereas index $j$ will loop over positive and negative samples taking into account their connectivity to $i$ captured by $\Delta W_{ij}$. We provide this expansion in the Suppl. Material.

%

\subsection{Geometric Interpretation.}
\label{sec:rel}

Below, we analyze  COLES through the lens of {\em Alignment and Uniformity on the Hypersphere} of  \citet{contr_align_uniform}. To this end, we decompose our objective into the so-called alignment and uniformity losses. Firstly,  \citet{mikolov2013distributed} have shown  that SampledNCE with the sigmoid non-linearity is a practical approximation of SoftMax contrastive loss, the latter suffering poor scalability \wrt the count of negative samples. For this reason, many contrastive approaches (DeepWalk, GraphSAGE, DGI, Graph2Gauss, \etc) adopt SampledNCE rather than SoftMax (GRACE) framework.

 Wang and Isola \cite{contr_align_uniform} have decomposed the SoftMax contrastive loss into $\mathcal{L}_{align}$ and $\mathcal{L}_{\text {umiform}}$ \cite{contr_align_uniform}:
\begin{equation}
\mathcal{L}(u,v, \mathcal{N})=\mathcal{L}_{\text{align}}(u,v) + \mathcal{L}'_{\text{uniform}}(u,v, \mathcal{N})=-\log \frac{e^{\mathbf{u}^\top \mathbf{v}}}{e^{\mathbf{u}^{\top} \mathbf{v}}+\sum_{u^{\prime} \in \mathcal{N}} e^{\mathbf{u}^{\prime\top}{\mathbf{v}}}},
\label{eq:arithm}
\end{equation}
where $\mathcal{N}$ is a sampled subset of negative samples, $u$ and $v$ are node indexes of so-called positive sample and anchor embeddings,  $\mathbf{u}$ and $\mathbf{v}$. Let $\langle\mathbf{u}, \mathbf{u}\rangle=\left\langle\mathbf{u}^{\prime}, \mathbf{u}^{\prime}\right\rangle=\langle\mathbf{v}, \mathbf{v}\rangle=\tau^{2}(\tau$ acts as the so-called temperature). Moreover, $\mathcal{L}_{\text {align }}=-\langle\mathbf{u}, \mathbf{v}\rangle$ and
$\mathcal{L}_{\text{uniform}}=\log \sum_{u^\ddagger\in\mathcal{N}\cup\{u\}} e^{\mathbf{u^\ddagger}^{\top}\mathbf{v}}$, that is, $\mathcal{L}_{\text {uniform }}$ is a logarithm of an arithmetic mean of RBF responses over the subset $\mathcal{N} \cup\{u\}$. Of course, computing the total loss $\mathcal{L}$ requires drawing $u$ and $v$ from the graph 
and summing over multiple $\mathcal{L}_{\text{align}}(u,v)$ and $\mathcal{L}'_{\text{uniform}}(u,v, \mathcal{N})$ but we skip this step and the argument variables of loss functions for brevity.

COLES can be decomposed into $\mathcal{L}_{align}$ and $\mathcal{L}_{\text {umiform}}$ \cite{contr_align_uniform} as follows:
\vspace{-0.2cm}
\begin{equation}
\mathcal{L}_{\text{align}} + \mathcal{L}'_{\text{uniform}}=-\log  e^{\mathbf{u}^{\top} \mathbf{v} }-\frac{1}{|\mathcal{N}|}\sum_{u'\in\mathcal{N}} \log e^{-\mathbf{u'}^{\top} \mathbf{v} } = 
-\log\frac{e^{\mathbf{u}^{\top} \mathbf{v}}}{\left(\Pi_{u'\in\mathcal{N}}e^{\mathbf{u'}^{\top} \mathbf{v}}\right)^{\frac{1}{|\mathcal{N}|}}},
\label{eq:geom}
\vspace{-0.1cm}
\end{equation}
where $\mathcal{L}_{\text {align }}$ remains the same with SoftMax but $\mathcal{L}'_{\text{uniform}}=\log\left(\Pi_{u'\in\mathcal{N}}e^{\mathbf{u'}^{\top} \mathbf{v}}\right)^{\frac{1}{|\mathcal{N}|}}$ is in fact a logarithm of the geometric mean of RBF responses over the subset $\mathcal{N}$. Thus, our loss can be seen as the ratio of geometric means over RBF functions.
Several authors (\eg, Gonzalez \cite{gonzalez2002digital}) noted that the geometric mean helps smooth out the Gaussian noise under the i.i.d. uniform sampling while loosing less information than the arithmetic mean. The geometric mean enjoys better confidence intervals the arithmetic mean given a small number of samples. As we sample few negative nodes for efficacy, we expect the geometric mean is more reliable. 
Eq. \eqref{eq:arithm} and \eqref{eq:geom} are just two specific cases of a generalized loss:
\vspace{-0.3cm}
\begin{equation}
    \mathcal{L}_{\text{align}} + \mathcal{L}''_{\text{uniform}}=-\log\frac{e^{\mathbf{u}^{\top} \mathbf{v}}}{M_p\left(
e^{\mathbf{u'_1}^{\top} \mathbf{v}},\cdots,e^{\mathbf{u'^{\top}_{|\mathcal{N}|}} \mathbf{v}}\right)},
\label{eq:gener}
\end{equation}
where $M_p(\cdot)$ in $\mathcal{L}''_{\text{uniform}}=\log{M_p\left(
e^{\mathbf{u'_1}^{\top} \mathbf{v}},\cdots,e^{\mathbf{u'^{\top}_{|\mathcal{N}|}} \mathbf{v}}\right)}$ is the so-called generalized mean. We introduce $M_p(\cdot)$ into the denominator  of Eq. \eqref{eq:gener} but it can be also introduced in the numerator. We investigate the geometric ($p\!=\!0$), arithmetic ($p\!=\!1$), harmonic ($p\!=\!-1$) and quadratic ($p\!=\!2$) means. 

\section{Related Works}

\vspace{-0.25cm}
\paragraph{Graph Embeddings.} 
%
Graph embedding methods such as Laplacian Eigenmaps~\cite{belkin2003laplacian} and IsoMap~\cite{tenenbaum2000global}  reduce the  dimensionality of representations by assuming the data lies on a low-dimensional manifold. 
With these methods, for a set of  high-dimensional data features, a similarity graph is built based on the pairwise feature similarity, and  each node embedded into a low-dimensional space. 
The graph is constructed from non-relational high dimensional data features, 
%
%
and Laplacian Eigenmaps ignore relations between dissimilar node pairs, that is, embeddings of dissimilar nodes are not penalized. 

To alleviate the above shortcomings, DeepWalk~\cite{perozzi2014deepwalk} uses truncated random walks to explore the network structure and utilizes the skip-gram model~\cite{mikolov2013efficient} for word embedding to derive the embedding vectors of nodes. LINE~\cite{tang2015line} explores a similar idea with an explicit objective function by setting the walk length as one and applying negative sampling~\cite{mikolov2013efficient}. 
Node2Vec~\cite{grover2016node2vec} interpolates  between breadth-  and depth-first sampling strategies to aggregate different types of neighborhoods.

\vspace{-0.25cm}
\paragraph{Representation Learning for Graph Neural Networks.}
Supervised and (semi-)supervised GNNs~\cite{kipf2016semi} require labeled datasets that may not be readily available. 
Yet, unsupervised GNNs have received little attention. 
GCN \cite{kipf2016semi} employs the minimization of reconstruction error as the objective function to train the encoder. 
GraphSAGE~\cite{hamilton2017inductive} incorporates objectives inspired by DeepWalk \eg, contrastive loss encouraging nearby nodes to have similar representations while preserving dissimilarity between representations of disparate nodes. 
%
DGI~\cite{velickovic2019deep}, inspired by Deep InfoMax (DIM)~\cite{hjelm2018learning}, proposes an objective with global-local sampling strategy, which maximizes the Mutual Information (MI) between global and local graph embeddings. 
In contrast, Augmented Multiscale Deep InfoMax (AMDIM)~\cite{bachman2019learning}  maximizes MI between multiple views of data. 
%
MVRLG~\cite{hassani2020contrastive}  contrasts encodings from first-order neighbors and a graph diffusion. MVRLG uses GCNs to learn node embeddings for different views. Fisher-Bures Adversary GCN \cite{uai_ke} assumes that the graph is  generated \wrt some observation noise. 
Graph-adaptive ReLU \cite{www_2022_grelu} uses an adaptive non-linearity in GCN. 
%
%
%
Multi-view  augmentation-based methods, not studied by us, are complementary to COLES. 
Moreover, linear networks \eg, SGC~\cite{wu2019simplifying} and S\textsuperscript{2}GC~\cite{zhu2021simple} capture the neighborhood and increasingly larger neighborhoods of each node, respectively. 
SGC and S\textsuperscript{2}GC have no projection layer, which results in embeddings of size equal to the input dimension. 
DGI \cite{velickovic2019deep}  uses the block-contrastive strategy \cite{saunshi2019theoretical} by treating negative samples as a difference of instances and  a summary of node embeddings for positive samples. 
Finally, COLES can  be  extended to other domains/problems \eg, time series/change point detection \cite{deldari2021time} or few-shot learning \cite{christian_subs,arl,Zhang_2020_ACCV}.

\vspace{-0.05cm}
\noindent\textbf{(Negative) Sampling.} 
Sampling node pairs  relies on random  walks~\cite{perozzi2014deepwalk} or second-order proximity \cite{tang2015line}, \etc. In contrast, COLES samples an  undirected graph based on the random graph sampling theory \cite{erdhos1959random}, where each edge is independently chosen with a prescribed probability $p'>0$.

\vspace{-0.25cm}
\section{Experiments}
We evaluate COLES on transductive and inductive node classification tasks. Node clustering is also evaluated. 
COLES is  compared to  state-of-the-art unsupervised, contrastive and (semi-)supervised methods. Unsupervised methods do not use label information except for the classifier. Contrastive methods use the contrastive setting to learn similarity/dissimilarity.  (Semi-)supervised methods use labels to train their projection layer and classifier. By {\em semi-supervised}, we mean that only a few of nodes used for training are  labeled. 
(Semi-)supervised models use a SoftMax classifier, whereas unsupervised and contrastive methods use a logistic regression classifier.

\noindent\textbf{Datasets.} 
\label{data:dataset}
COLES is evaluated on four citation networks: Cora, Citeseer, Pubmed, Cora Full~\cite{kipf2016semi,bojchevski2017deep} for transductive setting. We also employ the large scale Ogbn-arxiv from  OGB~\cite{hu2020open}. 
Finally, the  Reddit~\cite{zeng2019graphsaint} dataset is used in inductive setting. 
Table~\ref{tab:dataset} provides details of all datasets. 


\noindent\textbf{Metrics.}  Fixed data splits~\cite{yang2016revisiting} for transductive tasks are often used in evaluations between different models.
However, such an experimental setup may benefit easily overfitting models \cite{shchur2018pitfalls}. 
Thus, instead of fixed data splits, results are averaged over 50 random splits for each dataset and standard deviations are reported for empirical evaluation on transductive tasks. %
Moreover, we also test the performance under a different number of samples per class \ie, 5 and 20 samples per class. 
Typically, the performance for the inductive task is tested on relatively larger graphs. Thus, we choose fixed data splits as in previous papers~\cite{hamilton2017inductive,zeng2019graphsaint}, and we report the Micro-F1 scores averaged on 10 runs. 
%

\noindent\textbf{Baseline models.} 
We group baseline models into unsupervised, contrastive and (semi-)supervised methods, and implement them in the same framework/testbed. Contrastive methods include DeepWalk~\cite{perozzi2014deepwalk}, GCN+SampledNCE developed as an alternative to GraphSAGE+SampledNCE~\cite{hamilton2017inductive}, Graph2Gauss~\cite{bojchevski2017deep}, SCE~\cite{zhang2020sce}, DGI~\cite{velickovic2019deep}, GRACE~\cite{zhu2020deep}, GCA \cite{zhu2021graph_gca} and GraphCL \cite{you2020graph_graphcl}, which are our main competitors. Note that GRACE, GCA and GraphCL are based on multi-view and data augmentation, and GraphCL is mainly intended for graph classification. We do not study graph classification as it requires advanced node pooling \cite{deeper_look2} with mixed- or high-order statistics \cite{me_ATN,me_tensor_tech_rep,me_tensor}. 
We  compare results with  representative (semi-)supervised  GCN~\cite{kipf2016semi},  GAT~\cite{velickovic2019deep} and MixHop~\cite{abu2019mixhop} models. SGC and S\textsuperscript{2}GC are  unsupervised spectral filter networks. They do not have any learnable parameters that depend on labels, with exception of a classifier. 
To reduce the resulting dimensionality, we also add PCA-S\textsuperscript{2}GC and RP-S\textsuperscript{2}GC, which use PCA and random projections to obtain the projection layer on these methods. 
We extend our COLES framework with different GNNs: GCN, SGC and/or S\textsuperscript{2}GC, and we name them COLES-GCN, COLES-SGC and COLES-S\textsuperscript{2}GC. As COLES-GCN is a multi-layer non-linear encoder, the optimization of COLES-GCN is non-convex. The optimization of COLES-SGC and COLES-S\textsuperscript{2}GC  is convex if $\mL\!-\!\frac{\eta'}{\kappa}\sum_{k=1}^{\eta'}\mL_k^{(-)}\in\semipd{n}$, and COLES-GCN (Stiefel) is convex \wrt $\mP$.
We set hyperparameters based on the settings described in their papers. 







\noindent\textbf{General model setup.} 
For all  (semi-)supervised models, we use  early stopping on each random split and we capture the corresponding classification result.
For all  unsupervised models, we choose the embedding dimension to be 512 on Cora, Citeseer and Cora Full, and 256 on Pubmed. After the embeddings of nodes are learnt, a classifier is trained by applying the logistic regression in the embedding space. 
For inductive learning, methods based on COLES use 512-dimensional embeddings. Other hyperparameters for the baseline models are the same as  in original papers.

\noindent\textbf{Hyperparameter of our models.} 
In the transductive experiments, the detailed hyperparameter settings for Cora, Citeseer, Pubmed, and Cora Full are listed below.
For  COLES, we use the Adam optimizer with learning rates of $[0.001,0.0001,0.02,0.02]$ and the decay of $[5e\!-\!4, 1e\!-\!3, 5e\!-\!4, 2e\!-\!4]$. The number of training epochs are $[20,20,100,30]$, respectively. 
We sample 10 randomized adjacent matrices, and 5 negative samples for each node in each matrix on each dataset before training.
For the S\textsuperscript{2}GC and COLES-S\textsuperscript{2}GC, the number of propagation steps (layers) are 8 for all datasets except Cora Full (2 steps). For SGC and COLES-SGC, we use 2 steps for all datasets. 
%
%

\begin{table}[]
\vspace{-0.2cm}
\centering
\caption{The statistics of datasets.}
\resizebox{0.8\textwidth}{!}{ 
\begin{tabular}{llllll}
\toprule
Dataset   &Task & Nodes   & Edges      & Features & Classes \\
\midrule
Cora      &Transductive & 2,708   & 5,429      & 1,433    & 7       \\
Citeseer  &Transductive & 3,327   & 4,732      & 3,703    & 6       \\
Pubmed    &Transductive & 19,717  & 44,338     & 500      & 3       \\
Cora Full &Transductive & 19,793  & 65,311     & 8,710    & 70      \\
Ogbn-arxiv &Transductive & 169,343	& 1,166,243  & 128      & 40       \\
Reddit    &Inductive & 232,965 & 11,606,919 & 602      & 41      \\
\bottomrule
\end{tabular}
}
\label{tab:dataset}
\vspace{-0.3cm}
\end{table}


\begin{table*}[b]
\vspace{-0.5cm}
\hspace{-0.15cm}
  \centering
  \caption{Mean classification accuracy (\%) and the standard dev. over 50 random splits. 
  Numbers of labeled samples per class are  in parentheses. The best accuracy per column is in bold. 
  Models are organized into semi-supervised, contrastive and unsupervised groups. OOM means out of memory.
  }
  \resizebox{\textwidth}{!}{
  \fontsize{12}{12}\selectfont
    \begin{tabular}{clrrrrrrrr}
    \toprule
    \multicolumn{2}{c}{\multirow{2}[0]{*}{Method}}  & \multicolumn{2}{c}{Cora} & \multicolumn{2}{c}{Citeseer} & \multicolumn{2}{c}{Pubmed} & \multicolumn{2}{c}{Cora Full} \\
    \multicolumn{2}{c}{} & \multicolumn{1}{c}{(5)} & \multicolumn{1}{c}{(20)} & \multicolumn{1}{c}{(5)} & \multicolumn{1}{c}{(20)} & \multicolumn{1}{c}{(5)} & \multicolumn{1}{c}{(20)} & \multicolumn{1}{c}{(5)} & \multicolumn{1}{c}{(20)} \\
    \midrule
    \multirow{3}[0]{*}{$\substack{\text{ \fontsize{12}{12}\selectfont Semi-}\\\text{ \fontsize{12}{12}\selectfont supervised}}$ } & GCN   & 67.5$\pm$4.8 & 79.4$\pm$1.6 & 57.7$\pm$4.7 & 69.4$\pm$1.4 & 65.4$\pm$5.2 & 77.2$\pm$2.1 & 49.3$\pm$1.8 & 61.5$\pm$0.5 \\
          & GAT   & 71.2$\pm$3.5 & 79.6$\pm$1.5 & 54.9$\pm$5.0 & 69.1$\pm$1.5 & 65.5$\pm$4.6 & 75.4$\pm$2.3 & 43.9$\pm$1.5 & 56.9$\pm$0.6 \\
          & MixHop & 67.9$\pm$5.7 & 80.0$\pm$1.4 & 54.5$\pm$4.3 & 67.1$\pm$2.0 & 64.4$\pm$5.6 & 75.7$\pm$2.7 & 47.5$\pm$1.5 & 61.0$\pm$0.7 \\
    \midrule
    \multirow{11}[0]{*}{Contrastive} & DeepWalk & 60.3$\pm$4.0 &  70.5$\pm$1.9 &  38.3$\pm$2.9 &  45.6$\pm$2.0 &  60.3$\pm$5.6 &  70.8$\pm$2.6 &  38.9$\pm$1.4 & 51.1$\pm$0.7 \\\cdashline{2-2}[1pt/1pt]
          & GCN with  & \multirow{2}[0]{*}{61.3$\pm$4.3} & \multirow{2}[0]{*}{74.3$\pm$1.6} & \multirow{2}[0]{*}{42.3$\pm$3.4} & \multirow{2}[0]{*}{56.8$\pm$1.9} & \multirow{2}[0]{*}{60.9$\pm$5.7} & \multirow{2}[0]{*}{70.3$\pm$2.5} & \multirow{2}[0]{*}{32.7$\pm$1.9} & \multirow{2}[0]{*}{45.2$\pm$0.9} \\
          & SampledNCE$\!\!$          \\\cdashline{2-2}[1pt/1pt]
           & SAGE with  & \multirow{2}[0]{*}{65.0$\pm$3.5} & \multirow{2}[0]{*}{73.8$\pm$1.5} & \multirow{2}[0]{*}{48.0$\pm$3.5} & \multirow{2}[0]{*}{56.5$\pm$1.6} & \multirow{2}[0]{*}{64.1$\pm$6.1} & \multirow{2}[0]{*}{74.6$\pm$1.9} & \multirow{2}[0]{*}{35.0$\pm$1.4} & \multirow{2}[0]{*}{43.6$\pm$0.6} \\
          & SampledNCE$\!\!$\\\cdashline{2-2}[1pt/1pt]
          & Graph2Gauss   & 72.7$\pm$2.0 & 76.2$\pm$1.1 &  60.7$\pm$3.5 &  65.7$\pm$1.5 &  67.6$\pm$3.9 &  74.1$\pm$2.1 &  38.9$\pm$1.3 &  49.3$\pm$0.5 \\
          &  SCE & 74.3$\pm$2.7 & 80.2$\pm$1.1 & 65.4$\pm$2.9 & 70.7$\pm$1.2 & 65.7$\pm$6.0 & 75.8$\pm$2.2 & 50.7$\pm$1.5 & 60.6$\pm$0.6 \\
          & DGI   & 72.9$\pm$4.0 & 78.1$\pm$1.8 & 65.7$\pm$3.6 & 71.1$\pm$1.1 & 65.3$\pm$5.7 & 73.9$\pm$2.3 & 50.5$\pm$1.4 & 58.4$\pm$0.6 \\
          & {\CO COLES-GCN} & 73.8$\pm$3.4 &  80.8$\pm$1.3 &  66.0$\pm$2.6 &  69.0$\pm$1.3 &  62.7$\pm$4.6 &  72.7$\pm$2.1 &  47.3$\pm$1.5 & 58.9$\pm$0.5 \\\cdashline{2-2}[1pt/1pt]
          & {\CO COLES-GCN} & \multirow{2}[0]{*}{75.0$\pm$3.4} & \multirow{2}[0]{*}{81.0$\pm$1.3} & \multirow{2}[0]{*}{\textbf{67.9$\pm$2.3}} & \multirow{2}[0]{*}{\textbf{71.7$\pm$0.9}} & \multirow{2}[0]{*}{62.6$\pm$5.0} & \multirow{2}[0]{*}{73.2$\pm$2.6} & \multirow{2}[0]{*}{47.6$\pm$1.2} & \multirow{2}[0]{*}{59.2$\pm$0.5} \\
          & {\CO (Stiefel)}\\\cdashline{2-2}[1pt/1pt]
          & {\CO COLES-S\textsuperscript{2}GC} & \textbf{76.5$\pm$2.6} &  \textbf{81.5$\pm$1.2} &  67.5$\pm$2.2 &  71.3$\pm$1.0 &  66.0$\pm$5.2 &  \textbf{77.4$\pm$1.9} &  \textbf{50.8$\pm$1.4} & \textbf{61.8$\pm$0.5} \\
          \midrule
          \multirow{5}[0]{*}{$\substack{\text{ \fontsize{11.5}{11.5}\selectfont Contrastive +}\\\text{ \fontsize{11.5}{11.5}\selectfont Augmentation}}$} & GraphCL & 72.6$\pm$4.2 & 78.3$\pm$1.7 & 65.6$\pm$3.0 & 71.1$\pm$0.8 & OOM & OOM & OOM & OOM \\
          & GRACE & 64.9$\pm$4.2 & 73.9$\pm$1.6 & 61.8$\pm$3.9 & 68.4$\pm$1.6 & OOM & OOM & OOM & OOM \\
          & GCA & 61.5$\pm$4.9 & 75.8$\pm$1.9 & 43.2$\pm$3.6 & 55.7$\pm$1.9 & OOM & OOM & OOM & OOM \\\cdashline{2-2}[1pt/1pt]
          & {\CO COLES-GCN} & \multirow{2}[0]{*}{75.3$\pm$3.3} & \multirow{2}[0]{*}{81.0$\pm$1.3} & \multirow{2}[0]{*}{66.7$\pm$2.3} & \multirow{2}[0]{*}{69.8$\pm$1.3} & \multirow{2}[0]{*}{63.9$\pm$5.0} & \multirow{2}[0]{*}{73.4$\pm$2.5} & \multirow{2}[0]{*}{48.0$\pm$1.2} & \multirow{2}[0]{*}{59.4$\pm$0.5} \\
          & {\CO (+Aug)}\\     
      \midrule
      \multirow{4}[0]{*}{Unsupervised} & SGC   & 63.9$\pm$5.4 & 78.3$\pm$1.9 & 59.5$\pm$3.4 & 69.8$\pm$1.4 & 65.8$\pm$4.4 & 76.3$\pm$2.3 & 46.0$\pm$2.2 & 57.7$\pm$1.2 \\
          & S\textsuperscript{2}GC   & 71.4$\pm$4.4 & 81.3$\pm$1.2 & 60.3$\pm$4.0 & 69.5$\pm$1.2 & \textbf{67.6$\pm$4.2} & 73.3$\pm$2.0 & 41.8$\pm$1.7 & 60.0$\pm$0.5 \\
      & PCA-S\textsuperscript{2}GC & 72.1$\pm$3.8 &  81.2$\pm$1.3 &  61.0$\pm$3.5 &  68.8$\pm$1.3 &  67.5$\pm$4.3 &  73.2$\pm$2.0 &  42.3$\pm$1.7 & 59.3$\pm$0.6 \\
          & RP-S\textsuperscript{2}GC & 65.9$\pm$4.6 &  78.1$\pm$1.2 &  51.4$\pm$3.2 &  61.7$\pm$1.6 &  66.1$\pm$5.0 &  72.5$\pm$1.9 &  31.5$\pm$1.4 & 48.7$\pm$0.6 \\
          \bottomrule
    \end{tabular}%
    }
  \label{tab:trans}%
  \vspace{-0.1cm}
\end{table*}%

\subsection{Transductive Learning}
In this section, we consider transductive learning where all nodes are available in the training process.

\noindent\textbf{Contrastive Embedding Baselines \vs COLES.} 
 Table~\ref{tab:trans} shows that the performance of COLES-GCN  and the linear variant, COLES-S\textsuperscript{2}GC, are better than other unsupervised models. 
In particular, COLES-GCN  outperforms GCN+SampledNCE on all four datasets, which shows that   COLES has an advantage over the SampledNCE framework. 
In addition, COLES-S\textsuperscript{2}GC typically outperforms the best contrastive baseline DGI by up to 3.4\%. 
In Cora Full, we notice that S\textsuperscript{2}GC  underperforms when training with 5 samples. 
However, COLES-S\textsuperscript{2}GC is able to significantly boost its performance by 9\%. 
On  Citeseer with 5 training samples, COLES-S\textsuperscript{2}GC outperforms S\textsuperscript{2}GC by 6.8\%. We also note that COLES-GCN (Stiefel) outperforms COLES-GCN (based on the soft-orthogonality constraint) by up to 2.7\% but its performance below the performance of COLES-S\textsuperscript{2}GC.

Noteworthy is that for augmentation-based methods, COLES-GCN with augmentations denoted as  COLES-GCN (+Aug) outperforms  COLES-GCN without augmentations. COLES-GCN (+Aug) also outperforms GRACE and GCA, and GraphCL in most experiments. Nonetheless, COLES-S\textsuperscript{2}GC without any augmentations outperformed all augmentation-based methods.

Finally, Table \ref{tab:ogb} shows that COLES-S\textsuperscript{2}GC outperforms all other methods on the challenging  Ogbn-arxiv, while using a very small number of trainable parameters.

\noindent\textbf{Semi-supervised GNNs \vs COLES.}
Table~\ref{tab:trans} shows that the  contrastive GCN baselines perform worse than semi-supervised variants, especially when 20 labeled samples per class are available. 
In contrast, COLES-GCN outperformed the semi-supervised GCN on Cora by 6.3\% and 1.4\% given 5 and 20 labeled samples per class. COLES-GCN also outperforms GCN on Citeseer and Cora Full by 8.3\% and 6.3\% given 5 labeled samples per class. 
When the number of labels per class is 5, COLES-S\textsuperscript{2}GC outperforms GCN by a margin of 8.1\% on Cora and 9.4\% on Citeseer. 
These results show the superiority of COLES on four datasets when the number of samples per class is 5. 
Even for 20 labeled samples per class, COLES-S\textsuperscript{2}GC outperforms the best semi-supervised baselines on all four datasets \eg, by 1.7\% on Citeseer.
%
Semi-supervised models are  affected by the low number of labeled samples, which is consistent with \cite{li2018deeper}, \eg, for GAT and MixHop. 
The accuracy of COLES-GCN and COLES-S\textsuperscript{2}GC is not affected as significantly due to the contrastive setting. 

\noindent\textbf{Unsupervised GNNs \vs COLES.}
SGC and S\textsuperscript{2}GC are unsupervised LGNs as they are spectral filters which do not use labels (except for the classifier). 
%
Table~\ref{tab:trans} shows that COLES-S\textsuperscript{2}GC outperforms RP-S\textsuperscript{2}GC  and  PCA-S\textsuperscript{2}GC under the same size of projections. 
%
%
In most cases, COLES-S\textsuperscript{2}GC also outperforms the unsupervised S\textsuperscript{2}GC baseline (high-dimensional representation).

\begin{table}[t]
\vspace{-0.2cm}
\begin{minipage}[a]{0.48\linewidth}
  \centering
  \caption{The influence of the number ($\kappa$) of negative Laplacian graphs  on COLES-S\textsuperscript{2}GC. Parentheses show the no. of labeled samples p/class.}
  \label{tab:SSGCK}%
  \resizebox{\textwidth}{!}{  
    \begin{tabular}{llrrrr}
    \toprule
    &      & $\kappa\!\rightarrow\;$ 0     & 1     & 5     & 10 \\
    \midrule
    Cora& (20) & 79.88 & 81.43 & 81.18 & 81.17 \\
    Cora& (5) & 70.12 & 76.24 & 75.89 & 75.79 \\
    Citeseer& (20) & 69.42 & 70.71 & 70.61 & 70.61 \\
    Citeseer& (5) & 58.17 & 67.03 & 66.96 & 67.04 \\
    \bottomrule
    \end{tabular}}
    \end{minipage}
\hspace{0.2cm}
\begin{minipage}[a]{0.48\linewidth}
  \centering
  \caption{The influence of the number ($\kappa$) of negative Laplacian graphs on COLES-GCN. Parentheses show the no. of labeled samples p/class.}
  \label{tab:GCNK}%
  \resizebox{\textwidth}{!}{  
    \begin{tabular}{llrrrr}
    \toprule
    &     & $\kappa\!\rightarrow\;$ 0     & 1     & 5     & 10 \\
    \midrule
    Cora& (20) & 75.70 & 80.90 & 80.87 & 80.90 \\
    Cora& (5) & 60.97 & 74.14 & 74.11 & 74.07 \\
    Citeseer& (20) & 60.61 & 69.04 & 69.21 & 69.08 \\
    Citeseer& (5) & 45.31 & 65.85 & 66.08 & 66.01\\
    \bottomrule
    \end{tabular}}
    \end{minipage}
\end{table}%

\begin{table}[t]
    \vspace{-0.3cm}
    \centering
    \caption{The performance given various choices of the generalized mean $M_p$ for the uniformity loss.}
\label{tab:generalized}
    \resizebox{\textwidth}{!}{  
    \begin{tabular}{lllllllll}
    \hline
    \toprule
    \multicolumn{1}{c}{\multirow{2}[0]{*}{Method}}  & \multicolumn{2}{c}{Cora} & \multicolumn{2}{c}{Citeseer} & \multicolumn{2}{c}{Pubmed} & \multicolumn{2}{c}{Cora Full} \\
    \multicolumn{1}{c}{} & \multicolumn{1}{c}{(5)} & \multicolumn{1}{c}{(20)} & \multicolumn{1}{c}{(5)} & \multicolumn{1}{c}{(20)} & \multicolumn{1}{c}{(5)} & \multicolumn{1}{c}{(20)} & \multicolumn{1}{c}{(5)} & \multicolumn{1}{c}{(20)} \\
    \midrule
        Geometric ($M_0$) & \multirow{2}[0]{*}{76.5$\pm$2.6} & \multirow{2}[0]{*}{81.5$\pm$1.2} & \multirow{2}[0]{*}{67.5$\pm$2.2} & \multirow{2}[0]{*}{71.3$\pm$1.0} & \multirow{2}[0]{*}{66.0$\pm$5.2} & \multirow{2}[0]{*}{77.4$\pm$1.9} & \multirow{2}[0]{*}{50.8$\pm$1.4} & \multirow{2}[0]{*}{61.8$\pm$0.5} \\
        (COLES-S\textsuperscript{2}GC)\\\cdashline{1-1}[1pt/1pt]
        Arithmetic ($M_1$) & \multirow{2}[0]{*}{71.8$\pm$3.0}	& \multirow{2}[0]{*}{77.6$\pm$1.3}	& \multirow{2}[0]{*}{63.2$\pm$3.1}	& \multirow{2}[0]{*}{69.3$\pm$0.8}	& \multirow{2}[0]{*}{65.9$\pm$4.3}	& \multirow{2}[0]{*}{77.1$\pm$1.5}	& \multirow{2}[0]{*}{49.2$\pm$1.4}	& \multirow{2}[0]{*}{60.6$\pm$0.6} \\
        (SoftMax-Contrastive)\\\cdashline{1-1}[1pt/1pt]
        Harmonic ($M_{-1}$) & 75.2$\pm$3.5 & 80.7$\pm$1.2 & 64.7$\pm$2.4 & 70.9.$\pm$0.9 & 65.9$\pm$5.5 & 73.9$\pm$2.4 & 48.0$\pm$1.6 & 59.7$\pm$1.6 \\ 
        Quadratic ($M_2$)& 72.3$\pm$2.5	& 77.2$\pm$1.3	& 65.4$\pm$2.2	& 70.7.$\pm$0.8	& 65.6$\pm$4.5	& 77.3$\pm$1.5	& 49.2$\pm$1.5	& 60.6$\pm$1.6 \\ 
        \bottomrule
    \end{tabular}
    }
    \vspace{-0.2cm}
\end{table}

\begin{table}[b]
\vspace{-0.5cm}
\hspace{0.05cm}
\begin{minipage}[a]{0.48\linewidth}
\centering
\caption{Test Micro F1 Score (\%) averaged over 10 runs on Reddit. Results of other models are from original papers.}
\label{tab:reddit}
\resizebox{0.86\textwidth}{!}{  
\begin{tabular}{lll}
\toprule
Setting      & Model           & Test F1 \\
\midrule
             & SAGE-mean       & 95.0    \\
Supervised   & SAGE-LSTM       & \textbf{95.4}    \\
             & SAGE-GCN        & 93.0    \\
\midrule
\multirow{8}[0]{*}{Contrastive} & SAGE-mean       & 89.7    \\
             & SAGE-LSTM       & 90.7    \\
             & SAGE-GCN        & 90.8    \\
             & FastGCN         & 93.7    \\
             & DGI             & 94.0    \\
             & {\CO COLES-GCN}       &94.0 \\
             & {\CO COLES-SGC}       &94.8 \\
             & {\CO COLES-S\textsuperscript{2}GC}  & \textbf{95.4}\\
\bottomrule
\end{tabular}
}
\end{minipage}
\hspace{0.2cm}
\begin{minipage}[a]{0.48\linewidth}
\centering
\caption{Mean classification accuracy (\%) and the standard dev. over 10 runs on Ogbn-arxiv. Results of other models are from original papers.}
\label{tab:ogb}
\resizebox{1\textwidth}{!}{ 
\begin{tabular}{lcll}
\toprule
Method &   $\!\!\!\!$ & Test Acc.   & \#Params     \\
\midrule
MLP&         & 55.50$\pm$0.23 &  110,120     \\
Node2Vec&\kern-0.75cm\cite{grover2016node2vec}$\!\!\!\!$ & 70.07$\pm$0.13 &  21,818,792 \\
GraphZoom &\kern-0.75cm\cite{deng2020graphzoom}$\!\!\!\!$ & 71.18$\pm$0.18 &  8,963,624  \\
C\&S&\kern-0.75cm\cite{huang2021combining}$\!\!\!\!$ & 71.26$\pm$0.01 &  5,160       \\
SAGE-mean&\kern-0.75cm\cite{hamilton2017inductive}$\!\!\!\!$ & 71.49$\pm$0.27 &  218,664     \\
GCN&\kern-0.75cm\cite{kipf2016semi}$\!\!\!\!$ & 71.74$\pm$0.29 &  142,888     \\
DeeperGCN&\kern-0.75cm\cite{li2020deepergcn}$\!\!\!\!$ & 71.92$\pm$0.17 &  491,176     \\
SIGN&\kern-0.75cm\cite{rossi2020sign}$\!\!\!\!$& 71.95$\pm$0.11 &  3,566,128  \\
FrameLet&\kern-0.75cm\cite{zheng2021framelets}$\!\!\!\!$ & 71.97$\pm$0.12 &  1,633,183  \\
S\textsuperscript{2}GC&\kern-0.75cm\cite{zhu2021simple}$\!\!\!\!$ & 72.01$\pm$0.25 & 110,120\\
\midrule
{\CO COLES-S\textsuperscript{2}GC}   &  $\!\!\!\!$  & \textbf{72.48$\pm$0.25} &  110,120  \\
\bottomrule
\end{tabular}
}
\end{minipage}
\end{table}

\noindent\textbf{Negative Laplacian Eigenmaps.}
Below, we analyze how $\kappa$ in Eq.~\eqref{eq:CL} influences the performance. We set $\kappa\in\{0, 1,5,10\}$ on COLES-S\textsuperscript{2}GC and COLES-GCN given Cora and Citeseer with 5 and 20 labeled samples per class. The case of  $\kappa\!=\!0$ means no negative Laplacian Eigenmaps  are used, thus the solution simplifies to regular Laplacian Eigenmaps parametrized by GCN embeddings. 
%
Table~\ref{tab:SSGCK} shows that without the negative Laplacian Eigenmaps, the performance of COLES-S\textsuperscript{2}GC  drops significantly \ie, between  6\% and 9\% for 5 labeled samples per class. That means the negative  Laplacian Eigenmaps play important role which highlights the benefits of COLES. Although negative Laplacian Eigenmaps improve results, using $\kappa\!>\!1$ negative matrices improves the performance only marginally. 
Table~\ref{tab:GCNK} shows that COLES-GCN relies on negative Laplacian Eigenmaps. Without negative Laplacian Eigenmaps, the performance of COLES-GCN drops by 20\% on Citeseer with 5 samples per class. 
Even when 20 samples per class are used, if $\kappa\!=\!0$, the performance of COLES-GCN drops by 8.4\%. 

\begin{table*}[t]
\centering
\caption{The clustering performance on Cora, Citeseer and Pubmed.}
\label{tab:clustering}
\resizebox{\textwidth}{!}{  
\begin{tabular}{ll|lll|lll|lll}
\toprule
\multirow{2}[0]{*}{Method}          &  \multirow{2}[0]{*}{Input}        &       & Cora  &       &       & Citeseer &       &       & Pubmed &              \\
    &      & Acc\% & NMI\% & F1\%  & Acc\% & NMI\%    & F1\%  & Acc\% & NMI\%  & F1\%   \\
\midrule
k-means    & Feature & 34.65 & 16.73 & 25.42 & 38.49 & 17.02    & 30.47 & 57.32 & 29.12  & 57.35  \\
Spectral-f & Feature & 36.26 & 15.09 & 25.64 & 46.23 & 21.19    & 33.70 & 59.91 & 32.55  & 58.61  \\
Spectral-g & Graph   & 34.19 & 19.49 & 30.17 & 25.91 & 11.84    & 29.48 & 39.74 & 3.46   & 51.97  \\
DeepWalk   & Graph   & 46.74 & 31.75 & 38.06 & 36.15 & 9.66     & 26.70 & 61.86 & 16.71  & 47.06  \\
GAE        & Both    & 53.25 & 40.69 & 41.97 & 41.26 & 18.34    & 29.13 & 64.08 & 22.97  & 49.26  \\
VGAE       & Both    & 55.95 & 38.45 & 41.50 & 44.38 & 22.71    & 31.88 & 65.48 & 25.09  & 50.95 \\
ARGE       & Both    & 64.00 & 44.90 & 61.90 & 57.30 & 35.00    & 54.60 & 59.12 & 23.17  & 58.41  \\
ARVGE      & Both    & 62.66 & 45.28 & 62.15 & 54.40 & 26.10    & 52.90 & 58.22 & 20.62  & 23.04  \\
GCN      & Both    & 59.05 & 43.06 & 59.38 & 45.97 & 20.08    & 45.57 & 61.88 & 25.48  & 60.70  \\
SGC        & Both    &  62.87 & 50.05 &  58.60 & 52.77 & 32.90    & 63.90 & 69.09 & 31.64  & 68.45  \\
S\textsuperscript{2}GC  & Both    & 68.96 &  54.22 & \textbf{65.43} &  69.11 &  42.87&  64.65 &  68.18 & 31.82 &  67.81   \\
\midrule
{\CO COLES-GCN} & Both    & 60.74 &  45.49 & 59.33 &  63.28 & 37.54 & 59.17 &  63.46 & 25.73 & 63.42 \\\cdashline{1-1}[1pt/1pt]
{\CO COLES-GCN} & \multirow{2}[0]{*}{Both}    & \multirow{2}[0]{*}{62.46} &  \multirow{2}[0]{*}{47.01} & \multirow{2}[0]{*}{59.38} &  \multirow{2}[0]{*}{65.17} & \multirow{2}[0]{*}{38.90} & \multirow{2}[0]{*}{60.85} &  \multirow{2}[0]{*}{63.56} & \multirow{2}[0]{*}{25.81} & \multirow{2}[0]{*}{63.58} \\
{\CO (Stiefel)} & & & & & & & & & & \\\cdashline{1-1}[1pt/1pt]
{\CO COLES-SGC} & Both    & 65.62 &  52.32 & 56.95 &  68.24 & 43.09 & 63.85 &  \textbf{69.47} & 32.31 & \textbf{68.57} \\
{\CO COLES-S\textsuperscript{2}GC} & Both    & \textbf{69.70} &  \textbf{55.35} & 63.06 &  \textbf{69.20} & \textbf{44.41} &\textbf{64.70} &  68.76 & \textbf{33.42} &  68.12 \\
\bottomrule
\end{tabular}
}
\end{table*}

\vspace{-0.25cm}
\subsection{Uniformity Loss as the  Generalized Mean ($\mathbf{M_p}$).} Following the analysis presented in Section \ref{sec:rel}, Table \ref{tab:generalized} demonstrates the impact of the choices of the uniformity loss on the performance of COLES. To this end, we select the geometric ($M_0$), arithmetic ($M_1$), harmonic ($M_{-1}$) and quadratic($M_2$) means as examples of $M_p$ realizing the uniformity loss. On all the investigated datasets, the geometric mean  outperforms other variants.

\vspace{-0.25cm}
\subsection{Inductive Learning}
In inductive learning, models have no access to the test set, thus they need to  generalize well to unseen samples. 
Table~\ref{tab:reddit} shows that 
COLES enjoys a significant performance gain (1\% - 5\% in Micro-F1 scores), performing close to supervised methods with a low memory footprint. 
In contrast, DGI 
on Reddit triggers out-of-memory errors on Nvidia GTX 1080 GPU (94.0 Micro-F1 is taken from~\cite{velickovic2019deep}).

\subsection{Node Clustering}
%
We compare COLES-GCN and COLES-S\textsuperscript{2}GC with three types of clustering methods listed below:
\renewcommand{\labelenumi}{\roman{enumi}.}
\vspace{-0.25cm}
\hspace{-1.0cm}
\begin{enumerate}[leftmargin=0.6cm]
\item Methods that use only node features \eg, k-means and spectral clustering (spectral-f) construct a similarity matrix with the node features by a linear kernel.
\item Structural clustering methods that  only use  the graph structure: spectral clustering (spectral-g) that takes the graph adjacency matrix as the similarity matrix, and DeepWalk~\cite{perozzi2014deepwalk}.
\item Attributed graph clustering methods that use node features and the graph: Graph Autoencoder (GAE), Graph Variational Autoencoder (VGAE)~\cite{kipf2016semi}, Adversarially Regularized Graph Autoencoder (ARGE), Variational Graph Autoencoder (ARVGE)~\cite{pan2018adversarially}, SGC~\cite{wu2019simplifying} and S\textsuperscript{2}GC~\cite{zhu2021simple}.
\end{enumerate}
We measure the performance by the clustering Accuracy (Acc), Normalized Mutual Information (NMI) and macro F1-score (F1). We run each method 10 times on Cora, CiteSeer and PubMed. We report the clustering results in Table~\ref{tab:clustering}. 
We set the number of propagation steps to 8 for SGC, S\textsuperscript{2}GC, COLES-SGC and COLES-S\textsuperscript{2}GC, 
following the setting of \cite{zhang2019attributed}. 
We note that COLES-S\textsuperscript{2}GC outperforms S\textsuperscript{2}GC in most cases, whereas COLES-GCN  outperforms contrastive GCN on all datasets. 

\noindent\textbf{Scalability.} GraphSAGE and DGI 
require neighbor sampling which result in redundant forward/backward propagation steps (long runtime). 
%
In contrast, COLES-S\textsuperscript{2}GC enjoys a straightforward implementation which  reduces the memory usage and runtime significantly. 
For graphs with more than 100 thousands nodes and 10 millions edges (Reddit), our model  runs smoothly on NVIDIA 1080 GPU. 
Even on larger graph datasets, the closed-form solution is attractive as for COLES-S\textsuperscript{2}GC, the cost of eigen-decomposition depends on $d$ (a few of seconds on Reddit). 
%
The runtime of COLES-S\textsuperscript{2}GC is also favourable in comparison to multi-view augmentation-based GraphCL. Specifically, COLES-S\textsuperscript{2}GC took 0.3s, 1.4s, 7.3s and 16.4s on Cora, Citeseer, Pubmed and Cora Full, respectively.  GraphCL took 110.19s, 101.0s,  $\geq8$h and $\geq8$h respectively.

\section{Conclusions}
We have proposed a new network embedding, COnstrative Laplacian EigenmapS (COLES), which recognizes the  importance of negative sample pairs in Laplacian Eignemaps. Our COLES   works well with many backbones, \eg, COLES with GCN, SGC and S\textsuperscript{2}GC backbones outperforms many unsupervised, contrastive and (semi-)supervised methods. 
By applying the GAN-inspired analysis, we have shown that  SampledNCE  with the sigmoid non-linearity yields the JS divergence.  However, COLES uses the RBF non-linearity, which results in  the Kantorovich-Rubinstein duality;  COLES essentially minimizes a surrogate of Wasserstein distance, 
which offers a reasonable transportation plan, and helps avoid pitfalls of the JS divergence. 
%
%
%
%
Moreover, COLES takes advantage of the so-called block-contrastive loss whose family is known to perform better than their pair-wise contrastive counterparts. Cast as the alignment and uniformity losses, COLES  enjoys the more robust geometric mean rather than the arithmetic mean (used by SoftMax-Contrastive) as the uniformity loss. 

%
%

\ack
We would like to thank the reviewers for stimulating questions that helped us improve several aspects of our analysis. Hao Zhu is supported by an Australian Government Research Training Program (RTP) Scholarship. Ke Sun and Piotr Koniusz are supported by CSIRO’'s Machine Learning and 
Artificial Intelligence Future Science Platform (MLAI FSP).

\bibliography{main_arxiv}

\begin{thebibliography}{63}
\providecommand{\natexlab}[1]{#1}
\providecommand{\url}[1]{\texttt{#1}}
\expandafter\ifx\csname urlstyle\endcsname\relax
  \providecommand{\doi}[1]{doi: #1}\else
  \providecommand{\doi}{doi: \begingroup \urlstyle{rm}\Url}\fi

\bibitem[Abu-El-Haija et~al.(2019)Abu-El-Haija, Perozzi, Kapoor, Alipourfard,
  Lerman, Harutyunyan, Ver~Steeg, and Galstyan]{abu2019mixhop}
Sami Abu-El-Haija, Bryan Perozzi, Amol Kapoor, Nazanin Alipourfard, Kristina
  Lerman, Hrayr Harutyunyan, Greg Ver~Steeg, and Aram Galstyan.
\newblock Mixhop: Higher-order graph convolutional architectures via sparsified
  neighborhood mixing.
\newblock In \emph{International Conference on Machine Learning}, pages 21--29,
  2019.

\bibitem[Arjovsky et~al.(2017)Arjovsky, Chintala, and
  Bottou]{arjovsky2017wasserstein}
Martin Arjovsky, Soumith Chintala, and L{\'e}on Bottou.
\newblock Wasserstein generative adversarial networks.
\newblock In \emph{International Conference on Machine Learning}, pages
  214--223, 2017.

\bibitem[Arora et~al.(2019)Arora, Khandeparkar, Khodak, Plevrakis, and
  Saunshi]{saunshi2019theoretical}
Sanjeev Arora, Hrishikesh Khandeparkar, Mikhail Khodak, Orestis Plevrakis, and
  Nikunj Saunshi.
\newblock A theoretical analysis of contrastive unsupervised representation
  learning.
\newblock In \emph{International Conference on Machine Learning}, pages
  5628--5637, 2019.

\bibitem[Bachman et~al.(2019)Bachman, Hjelm, and
  Buchwalter]{bachman2019learning}
Philip Bachman, R~Devon Hjelm, and William Buchwalter.
\newblock Learning representations by maximizing mutual information across
  views.
\newblock \emph{arXiv preprint arXiv:1906.00910}, 2019.

\bibitem[Belkin and Niyogi(2003)]{belkin2003laplacian}
Mikhail Belkin and Partha Niyogi.
\newblock Laplacian eigenmaps for dimensionality reduction and data
  representation.
\newblock \emph{Neural computation}, 15\penalty0 (6):\penalty0 1373--1396,
  2003.

\bibitem[Bojchevski and G{\"u}nnemann(2017)]{bojchevski2017deep}
Aleksandar Bojchevski and Stephan G{\"u}nnemann.
\newblock Deep gaussian embedding of graphs: Unsupervised inductive learning
  via ranking.
\newblock \emph{arXiv preprint arXiv:1707.03815}, 2017.

\bibitem[Cai et~al.(2018)Cai, Zheng, and Chang]{cai2018comprehensive}
Hongyun Cai, Vincent~W Zheng, and Kevin Chen-Chuan Chang.
\newblock A comprehensive survey of graph embedding: Problems, techniques, and
  applications.
\newblock \emph{IEEE Transactions on Knowledge and Data Engineering},
  30\penalty0 (9):\penalty0 1616--1637, 2018.

\bibitem[Carreira-Perpi\~{n}an(2010)]{elastic_net}
Miguel~\'{A}. Carreira-Perpi\~{n}an.
\newblock The elastic embedding algorithm for dimensionality reduction.
\newblock In \emph{International Conference on Machine Learning}, page
  167–174, 2010.

\bibitem[Deldari et~al.(2021)Deldari, Smith, Xue, and Salim]{deldari2021time}
Shohreh Deldari, Daniel~V Smith, Hao Xue, and Flora~D Salim.
\newblock Time series change point detection with self-supervised contrastive
  predictive coding.
\newblock In \emph{Proceedings of the Web Conference}, pages 3124--3135, 2021.

\bibitem[Deng et~al.(2020)Deng, Zhao, Wang, Zhang, and Feng]{deng2020graphzoom}
C~Deng, Z~Zhao, Y~Wang, Z~Zhang, and Z~Feng.
\newblock Graphzoom: A multi-level spectral approach for accurate and scalable
  graph embedding.
\newblock In \emph{International Conference on Learning Representations}, 2020.

\bibitem[ErdHos and R{\'e}nyi(1959)]{erdhos1959random}
Paul ErdHos and Alfr{\'e}d R{\'e}nyi.
\newblock On random graphs.
\newblock \emph{Publicationes Mathematicae}, 6:\penalty0 290--297, 1959.

\bibitem[Gonzalez et~al.(2002)Gonzalez, Woods, et~al.]{gonzalez2002digital}
Rafael~C Gonzalez, Richard~E Woods, et~al.
\newblock Digital image processing, 2002.

\bibitem[Grover and Leskovec(2016)]{grover2016node2vec}
Aditya Grover and Jure Leskovec.
\newblock node2vec: Scalable feature learning for networks.
\newblock In \emph{ACM SIGKDD international conference on Knowledge discovery
  and Data Mining}, pages 855--864, 2016.

\bibitem[Gutmann and Hyv{\"a}rinen(2010)]{gutmann2010noise}
Michael Gutmann and Aapo Hyv{\"a}rinen.
\newblock Noise-contrastive estimation: A new estimation principle for
  unnormalized statistical models.
\newblock In \emph{Proceedings of the thirteenth international conference on
  artificial intelligence and statistics}, pages 297--304. JMLR Workshop and
  Conference Proceedings, 2010.

\bibitem[Hamilton et~al.(2017)Hamilton, Ying, and
  Leskovec]{hamilton2017inductive}
Will Hamilton, Zhitao Ying, and Jure Leskovec.
\newblock Inductive representation learning on large graphs.
\newblock In \emph{Advances in Neural Information Processing Systems}, pages
  1024--1034, 2017.

\bibitem[Hassani and Khasahmadi(2020)]{hassani2020contrastive}
Kaveh Hassani and Amir~Hosein Khasahmadi.
\newblock Contrastive multi-view representation learning on graphs.
\newblock In \emph{International Conference on Machine Learning}, pages
  4116--4126, 2020.

\bibitem[He and Niyogi(2004)]{he2004locality}
Xiaofei He and Partha Niyogi.
\newblock Locality preserving projections.
\newblock \emph{Advances in Neural Information Processing Systems}, 16\penalty0
  (16):\penalty0 153--160, 2004.

\bibitem[Hjelm et~al.(2018)Hjelm, Fedorov, Lavoie-Marchildon, Grewal, Bachman,
  Trischler, and Bengio]{hjelm2018learning}
R~Devon Hjelm, Alex Fedorov, Samuel Lavoie-Marchildon, Karan Grewal, Phil
  Bachman, Adam Trischler, and Yoshua Bengio.
\newblock Learning deep representations by mutual information estimation and
  maximization.
\newblock \emph{arXiv preprint arXiv:1808.06670}, 2018.

\bibitem[Hu et~al.(2020)Hu, Fey, Zitnik, Dong, Ren, Liu, Catasta, and
  Leskovec]{hu2020open}
Weihua Hu, Matthias Fey, Marinka Zitnik, Yuxiao Dong, Hongyu Ren, Bowen Liu,
  Michele Catasta, and Jure Leskovec.
\newblock Open graph benchmark: Datasets for machine learning on graphs.
\newblock \emph{arXiv preprint arXiv:2005.00687}, 2020.

\bibitem[Huang et~al.(2021)Huang, He, Singh, Lim, and
  Benson]{huang2021combining}
Qian Huang, Horace He, Abhay Singh, Ser-Nam Lim, and Austin Benson.
\newblock Combining label propagation and simple models out-performs graph
  neural networks.
\newblock In \emph{International Conference on Learning Representations}, 2021.

\bibitem[Kingma and Ba(2014)]{kingma2014adam}
Diederik~P Kingma and Jimmy Ba.
\newblock Adam: A method for stochastic optimization.
\newblock \emph{arXiv preprint arXiv:1412.6980}, 2014.

\bibitem[Kipf and Welling(2016)]{kipf2016semi}
Thomas~N Kipf and Max Welling.
\newblock Semi-supervised classification with graph convolutional networks.
\newblock \emph{arXiv preprint arXiv:1609.02907}, 2016.

\bibitem[Klicpera et~al.(2019)Klicpera, Bojchevski, and
  G{\"u}nnemann]{klicpera2018predict}
Johannes Klicpera, Aleksandar Bojchevski, and Stephan G{\"u}nnemann.
\newblock Predict then propagate: Graph neural networks meet personalized
  pagerank.
\newblock In \emph{International Conference on Learning Representations}, 2019.

\bibitem[Koniusz and Zhang(2020)]{deeper_look2}
Piotr Koniusz and Hongguang Zhang.
\newblock Power normalizations in fine-grained image, few-shot image and graph
  classification.
\newblock In \emph{IEEE Transactions on Pattern Analysis and Machine
  Intelligence}, 2020.
\newblock \doi{10.1109/TPAMI.2021.3107164}.

\bibitem[Koniusz et~al.(2013{\natexlab{a}})Koniusz, Yan, Gosselin, and
  Mikolajczyk]{me_tensor_tech_rep}
Piotr Koniusz, Fei Yan, Philippe-Henri Gosselin, and Krystian Mikolajczyk.
\newblock {Higher-order Occurrence Pooling on Mid- and Low-level Features:
  Visual Concept Detection}.
\newblock Technical report, INRIA, September 2013{\natexlab{a}}.
\newblock URL \url{https://hal.inria.fr/hal-00922524}.

\bibitem[Koniusz et~al.(2013{\natexlab{b}})Koniusz, Yan, and
  Mikolajczyk]{me_ATN}
Piotr Koniusz, Fei Yan, and Krystian Mikolajczyk.
\newblock Comparison of mid-level feature coding approaches and pooling
  strategies in visual concept detection.
\newblock \emph{Computer Vision and Image Understanding}, 117\penalty0
  (5):\penalty0 479 -- 492, 2013{\natexlab{b}}.
\newblock ISSN 1077-3142.

\bibitem[Koniusz et~al.(2016)Koniusz, Yan, Gosselin, and
  Mikolajczyk]{me_tensor}
Piotr Koniusz, Fei Yan, Philippe-Henri Gosselin, and Krystian Mikolajczyk.
\newblock Higher-order occurrence pooling for bags-of-words: {Visual} concept
  detection.
\newblock \emph{IEEE Transactions on Pattern Analysis and Machine
  Intelligence}, 2016.

\bibitem[Levy and Goldberg(2014)]{levy2014neural}
Omer Levy and Yoav Goldberg.
\newblock Neural word embedding as implicit matrix factorization.
\newblock \emph{Advances in Neural Information Processing Systems},
  27:\penalty0 2177--2185, 2014.

\bibitem[Lezcano~Casado(2019)]{NEURIPS2019_stiefel}
Mario Lezcano~Casado.
\newblock Trivializations for gradient-based optimization on manifolds.
\newblock In H.~Wallach, H.~Larochelle, A.~Beygelzimer, F.~d\textquotesingle
  Alch\'{e}-Buc, E.~Fox, and R.~Garnett, editors, \emph{Advances in Neural
  Information Processing Systems}, volume~32, 2019.

\bibitem[Li et~al.(2020)Li, Xiong, Thabet, and Ghanem]{li2020deepergcn}
Guohao Li, Chenxin Xiong, Ali Thabet, and Bernard Ghanem.
\newblock Deepergcn: All you need to train deeper gcns.
\newblock \emph{arXiv preprint arXiv:2006.07739}, 2020.

\bibitem[Li et~al.(2018)Li, Han, and Wu]{li2018deeper}
Qimai Li, Zhichao Han, and Xiao-Ming Wu.
\newblock Deeper insights into graph convolutional networks for semi-supervised
  learning.
\newblock In \emph{AAAI Conference on Artificial Intelligence}, pages
  3538--3545, 2018.

\bibitem[Mikolov et~al.(2013{\natexlab{a}})Mikolov, Chen, Corrado, and
  Dean]{mikolov2013efficient}
Tomas Mikolov, Kai Chen, Greg Corrado, and Jeffrey Dean.
\newblock Efficient estimation of word representations in vector space.
\newblock \emph{arXiv preprint arXiv:1301.3781}, 2013{\natexlab{a}}.

\bibitem[Mikolov et~al.(2013{\natexlab{b}})Mikolov, Sutskever, Chen, Corrado,
  and Dean]{mikolov2013distributed}
Tomas Mikolov, Ilya Sutskever, Kai Chen, Gregory~S Corrado, and Jeffrey Dean.
\newblock Distributed representations of words and phrases and their
  compositionality.
\newblock In \emph{Advances in Neural Information Processing Systems},
  2013{\natexlab{b}}.

\bibitem[Pan et~al.(2018)Pan, Hu, Long, Jiang, Yao, and
  Zhang]{pan2018adversarially}
S~Pan, R~Hu, G~Long, J~Jiang, L~Yao, and C~Zhang.
\newblock Adversarially regularized graph autoencoder for graph embedding.
\newblock In \emph{International Joint Conference on Artificial Intelligence},
  2018.

\bibitem[Perozzi et~al.(2014)Perozzi, Al-Rfou, and Skiena]{perozzi2014deepwalk}
Bryan Perozzi, Rami Al-Rfou, and Steven Skiena.
\newblock Deepwalk: Online learning of social representations.
\newblock In \emph{Proceedings of the 20th ACM SIGKDD international conference
  on Knowledge discovery and data mining}, pages 701--710, 2014.

\bibitem[Ramirez et~al.(2010)Ramirez, Sprechmann, and Sapiro]{incoherence}
Ignacio Ramirez, Pablo Sprechmann, and Guillermo Sapiro.
\newblock Classification and clustering via dictionary learning with structured
  incoherence and shared features.
\newblock In \emph{IEEE Conference on Computer Vision and Pattern Recognition},
  pages 3501--3508, 2010.
\newblock \doi{10.1109/CVPR.2010.5539964}.

\bibitem[Rossi et~al.(2020)Rossi, Frasca, Chamberlain, Eynard, Bronstein, and
  Monti]{rossi2020sign}
Emanuele Rossi, Fabrizio Frasca, Ben Chamberlain, Davide Eynard, Michael
  Bronstein, and Federico Monti.
\newblock Sign: Scalable inception graph neural networks.
\newblock \emph{arXiv preprint arXiv:2004.11198}, 2020.

\bibitem[Shchur et~al.(2018)Shchur, Mumme, Bojchevski, and
  G{\"u}nnemann]{shchur2018pitfalls}
Oleksandr Shchur, Maximilian Mumme, Aleksandar Bojchevski, and Stephan
  G{\"u}nnemann.
\newblock Pitfalls of graph neural network evaluation.
\newblock \emph{arXiv preprint arXiv:1811.05868}, 2018.

\bibitem[Simon et~al.(2020)Simon, Koniusz, Nock, and Harandi]{christian_subs}
Christian Simon, Piotr Koniusz, Richard Nock, and Mehrtash Harandi.
\newblock Adaptive subspaces for few-shot learning.
\newblock In \emph{IEEE Conference on Computer Vision and Pattern Recognition},
  2020.

\bibitem[Sun et~al.(2019)Sun, Koniusz, and Wang]{uai_ke}
Ke~Sun, Piotr Koniusz, and Zhen Wang.
\newblock Fisher-bures adversary graph convolutional networks.
\newblock \emph{Conference on Uncertainty in Artificial Intelligence},
  115:\penalty0 465--475, 2019.

\bibitem[Tang et~al.(2015)Tang, Qu, Wang, Zhang, Yan, and Mei]{tang2015line}
Jian Tang, Meng Qu, Mingzhe Wang, Ming Zhang, Jun Yan, and Qiaozhu Mei.
\newblock Line: Large-scale information network embedding.
\newblock In \emph{International Conference on World Wide Web}, pages
  1067--1077, 2015.

\bibitem[Tenenbaum et~al.(2000)Tenenbaum, De~Silva, and
  Langford]{tenenbaum2000global}
Joshua~B Tenenbaum, Vin De~Silva, and John~C Langford.
\newblock A global geometric framework for nonlinear dimensionality reduction.
\newblock \emph{Science}, 290\penalty0 (5500):\penalty0 2319--2323, 2000.

\bibitem[Veli{\v{c}}kovi{\'c} et~al.(2017)Veli{\v{c}}kovi{\'c}, Cucurull,
  Casanova, Romero, Lio, and Bengio]{velivckovic2017graph}
Petar Veli{\v{c}}kovi{\'c}, Guillem Cucurull, Arantxa Casanova, Adriana Romero,
  Pietro Lio, and Yoshua Bengio.
\newblock Graph attention networks.
\newblock \emph{arXiv preprint arXiv:1710.10903}, 2017.

\bibitem[Velickovic et~al.(2019)Velickovic, Fedus, Hamilton, Lio, Bengio, and
  Hjelm]{velickovic2019deep}
Petar Velickovic, William Fedus, William~L Hamilton, Pietro Lio, Yoshua Bengio,
  and R~Devon Hjelm.
\newblock Deep graph infomax.
\newblock In \emph{International Conference on Learning Representations}, 2019.

\bibitem[Villani(2009)]{ot}
C\'edric Villani.
\newblock \emph{Optimal {T}ransport, {O}ld and {N}ew}.
\newblock Springer-Verlag Berlin Heidelberg, 2009.

\bibitem[Wang et~al.(2020)Wang, Shen, Huang, Wu, Dong, and
  Kanakia]{wang2020microsoft}
Kuansan Wang, Zhihong Shen, Chiyuan Huang, Chieh-Han Wu, Yuxiao Dong, and
  Anshul Kanakia.
\newblock Microsoft academic graph: When experts are not enough.
\newblock \emph{Quantitative Science Studies}, 1\penalty0 (1):\penalty0
  396--413, 2020.

\bibitem[Wang and Isola(2020)]{contr_align_uniform}
Tongzhou Wang and Phillip Isola.
\newblock Understanding contrastive representation learning through alignment
  and uniformity on the hypersphere.
\newblock In \emph{International Conference on Machine Learning}, volume 119,
  pages 9929--9939, 2020.

\bibitem[Weng(2019)]{weng2019gan}
Lilian Weng.
\newblock From {GAN} to {WGAN}.
\newblock \emph{arXiv preprint arXiv:1904.08994}, 2019.

\bibitem[Wu et~al.(2019)Wu, Zhang, Souza~Jr, Fifty, Yu, and
  Weinberger]{wu2019simplifying}
Felix Wu, Tianyi Zhang, Amauri Holanda~de Souza~Jr, Christopher Fifty, Tao Yu,
  and Kilian~Q Weinberger.
\newblock Simplifying graph convolutional networks.
\newblock \emph{arXiv preprint arXiv:1902.07153}, 2019.

\bibitem[Yang et~al.(2020)Yang, Ding, Zhou, Yang, Zhou, and
  Tang]{yang2020understanding}
Zhen Yang, Ming Ding, Chang Zhou, Hongxia Yang, Jingren Zhou, and Jie Tang.
\newblock Understanding negative sampling in graph representation learning.
\newblock In \emph{ACM SIGKDD International Conference on Knowledge Discovery
  and Data Mining}, pages 1666--1676, 2020.

\bibitem[Yang et~al.(2016)Yang, Cohen, and Salakhudinov]{yang2016revisiting}
Zhilin Yang, William Cohen, and Ruslan Salakhudinov.
\newblock Revisiting semi-supervised learning with graph embeddings.
\newblock In \emph{International Conference on Machine Learning}, pages 40--48,
  2016.

\bibitem[You et~al.(2020)You, Chen, Sui, Chen, Wang, and
  Shen]{you2020graph_graphcl}
Yuning You, Tianlong Chen, Yongduo Sui, Ting Chen, Zhangyang Wang, and Yang
  Shen.
\newblock Graph contrastive learning with augmentations.
\newblock \emph{Advances in Neural Information Processing Systems},
  33:\penalty0 5812--5823, 2020.

\bibitem[Zeng et~al.(2019)Zeng, Zhou, Srivastava, Kannan, and
  Prasanna]{zeng2019graphsaint}
Hanqing Zeng, Hongkuan Zhou, Ajitesh Srivastava, Rajgopal Kannan, and Viktor
  Prasanna.
\newblock Graphsaint: Graph sampling based inductive learning method.
\newblock \emph{arXiv preprint arXiv:1907.04931}, 2019.

\bibitem[Zhang et~al.(2021)Zhang, Koniusz, Jian, Li, and Torr]{arl}
Hongguang Zhang, Piotr Koniusz, Songlei Jian, Hongdong Li, and Philip H.~S.
  Torr.
\newblock Rethinking class relations: Absolute-relative supervised and
  unsupervised few-shot learning.
\newblock In \emph{IEEE Conference on Computer Vision and Pattern Recognition},
  pages 9432--9441, 2021.

\bibitem[Zhang et~al.(2020{\natexlab{a}})Zhang, Luo, Wang, and
  Koniusz]{Zhang_2020_ACCV}
Shan Zhang, Dawei Luo, Lei Wang, and Piotr Koniusz.
\newblock Few-shot object detection by second-order pooling.
\newblock In \emph{Asian Conference on Computer Vision}, 2020{\natexlab{a}}.

\bibitem[Zhang et~al.(2020{\natexlab{b}})Zhang, Huang, Zhou, and
  Zhou]{zhang2020sce}
Shengzhong Zhang, Zengfeng Huang, Haicang Zhou, and Ziang Zhou.
\newblock Sce: Scalable network embedding from sparsest cut.
\newblock In \emph{ACM SIGKDD International Conference on Knowledge Discovery
  and Data Mining}, pages 257--265, 2020{\natexlab{b}}.

\bibitem[Zhang et~al.(2019)Zhang, Liu, Li, and Wu]{zhang2019attributed}
Xiaotong Zhang, Han Liu, Qimai Li, and Xiao-Ming Wu.
\newblock Attributed graph clustering via adaptive graph convolution.
\newblock \emph{arXiv preprint arXiv:1906.01210}, 2019.

\bibitem[Zhang et~al.(2022)Zhang, Zhu, Meng, Koniusz, and King]{www_2022_grelu}
Yifei Zhang, Hao Zhu, Ziqiao Meng, Piotr Koniusz, and Irwin King.
\newblock Graph-adaptive rectified linear unit for graph neural networks.
\newblock In \emph{Proceedings of the Web Conference}, 2022.

\bibitem[Zheng et~al.(2021)Zheng, Zhou, Gao, Wang, Li{\'o}, Li, and
  Montufar]{zheng2021framelets}
Xuebin Zheng, Bingxin Zhou, Junbin Gao, Yuguang Wang, Pietro Li{\'o}, Ming Li,
  and Guido Montufar.
\newblock How framelets enhance graph neural networks.
\newblock In Marina Meila and Tong Zhang, editors, \emph{International
  Conference on Machine Learning}, volume 139, pages 12761--12771, 2021.

\bibitem[Zhu and Koniusz(2021{\natexlab{a}})]{zhu2021refine}
Hao Zhu and Piotr Koniusz.
\newblock Refine: Random range finder for network embedding.
\newblock In \emph{ACM Conference on Information and Knowledge Management},
  2021{\natexlab{a}}.

\bibitem[Zhu and Koniusz(2021{\natexlab{b}})]{zhu2021simple}
Hao Zhu and Piotr Koniusz.
\newblock Simple spectral graph convolution.
\newblock In \emph{International Conference on Learning Representations},
  2021{\natexlab{b}}.

\bibitem[Zhu et~al.(2020)Zhu, Xu, Yu, Liu, Wu, and Wang]{zhu2020deep}
Yanqiao Zhu, Yichen Xu, Feng Yu, Qiang Liu, Shu Wu, and Liang Wang.
\newblock Deep graph contrastive representation learning.
\newblock \emph{arXiv preprint arXiv:2006.04131}, 2020.

\bibitem[Zhu et~al.(2021)Zhu, Xu, Yu, Liu, Wu, and Wang]{zhu2021graph_gca}
Yanqiao Zhu, Yichen Xu, Feng Yu, Qiang Liu, Shu Wu, and Liang Wang.
\newblock Graph contrastive learning with adaptive augmentation.
\newblock In \emph{Proceedings of the Web Conference}, pages 2069--2080, 2021.

\end{thebibliography}
\bibliographystyle{plainnat}

\newpage
\appendix
\title{Contrastive  Laplacian Eigenmaps\\(Supplementary Material)}

\author{
Hao Zhu$^{\dagger, \S}\quad$ Ke Sun$^{\S, \dagger}\quad$ Piotr Koniusz\textsuperscript{ \textasteriskcentered}$^{,\S, \dagger}$\\
$^{\S}$Data61/CSIRO $\;\;^{\dagger}$Australian National University\\
allenhaozhu@gmail.com, sunk@ieee.org, piotr.koniusz@data61.csiro.au
}

\maketitle

\section{Derivations of Contrastive Laplacian Eigenmaps}
%
In this section, we perform the transition of Eq.~\eqref{eq:final1} into Eq.\eqref{eq:CL}. We note that Eq.~\eqref{eq:final1} relies on two terms: $\mathbb{E}_{v \sim p_{d}(v)}\left[\mathbb{E}_{u \sim p_{d}(u \mid v)} (\mathbf{u}^{\top} \mathbf{v})\right]$ and $\eta\mathbb{E}_{v \sim p_{d}(v)}\left[\mathbb{E}_{u'\sim p_{n}\left(u'\mid v\right)} \left(-\mathbf{u}'^\top\mathbf{v}\right)\right]$.
The above two terms are evaluated over two different distributions  $u \sim p_{d}(u \mid v)$ and $u'\sim p_{n}\left(u'\mid v\right)$, respectively. 
Below we discuss how to reformulate $\mathbb{E}_{v \sim p_{d}(v)}[\mathbb{E}_{u \sim p_{d}(u \mid v)}\mathbf{u}^\top\mathbf{v}]$ 
into the matrix form $\mathbf{\mathbf{Y}^\top\mathbf{L}\mathbf{Y}}$ (reformulation of the second term can be performed by analogy), where $\mathbf{L}\!=\!\mathbf{I}\!-\!\mathbf{W}^{(+)}$ and  $\mathbf{W}^{(+)}=\mathbf{D}^{-1/2}\widehat{\mathbf{W}}\mathbf{D}^{-1/2}$ Let $p_{d}(v) = \frac{1}{\sqrt{D_{vv}}}$ and $p_{d}(u \mid v) = \frac{{W}_{uv}}{\sqrt{D_{uu}}}$. Then we have:
\comment{
\begin{equation}
\mathbb{E}_{v \sim p_{d}(v)}[\mathbb{E}_{u \sim p_{d}(u \mid v)} \mathbf{u}^\top\mathbf{v}] = \sum_{u, v} \frac{\widehat{W}_{uv}}{\sqrt{D_{vv}D_{uu}}}\mathbf{u}^\top\mathbf{v}.
\end{equation}
}
\begin{align}
    \mathbb{E}_{v \sim p_{d}(v)}[\mathbb{E}_{u \sim p_{d}(u \mid v)}\mathbf{u}^\top\mathbf{v}] &=     \sum_{u, v} \frac{\widehat{W}_{uv}}{\sqrt{D_{vv}D_{uu}}}\mathbf{u}^\top\mathbf{v} 
    =\sum_i^{d'} \sum_{u, v} \frac{\widehat{W}_{uv}}{\sqrt{D_{vv}D_{uu}}}u_i v_i.
    \label{eq:pos1}
\end{align}
Note our slight abuse of notations where $u_i$ and $v_i$ are $i$-th coefficients of vectors $\mathbf{u}$ and $\mathbf{v}$, whereas $u$ and $v$ are note indexes. 
We notice that $\sum_{u, v} \frac{\widehat{W}_{uv}}{\sqrt{D_{vv}D_{uu}}}u_i v_i$ has a bilinear form $\langle\mathbf{Y}_i,\mathbf{W}^{(+)}\mathbf{Y}_i\rangle$, which leads to:
\begin{equation}
    \sum_i^{d'}\mathbf{y}_i^\top\mathbf{W}^{(+)}\mathbf{y}_i = \trace(\mathbf{Y}^\top\mathbf{W}^{(+)}\mathbf{Y}),
\label{eq:pos}
\end{equation}
where $\mathbf{u}$ and $\mathbf{v}$ are rows of $\mathbf{Y}$. Moreover, $u_i$ denotes the $i$-th element of the vector $u$ and $\mathbf{y}_i$ is the $i$-th column of the matrix $\mathbf{Y}$.

By analogy, if we sample $\kappa$ 
for the random graph (negative graph), we have:
\begin{equation}
    \mathbb{E}_{v \sim p_{d}(v)}[\mathbb{E}_{u' \sim p_{n}(u' \mid v)} -\mathbf{u'}^\top\mathbf{v}] =
    -\frac{1}{\kappa}\sum_{k=1}^\kappa\mathbb{E}_{v \sim p_{d}(v)}[\mathbb{E}_{u' \sim p^*_{n}(u' \mid v)} \mathbf{u'}^\top\mathbf{v}] =-\frac{1}{\kappa}\sum_{k=1}^\kappa
    \trace(\mathbf{Y}^\top\mathbf{W}_k^{(-)}\mathbf{Y}),
    \label{eq:neg}
\end{equation}
where $p^*_n(u'|v)$ represents some uniform probability $p'>0$ of creating the negative links between nodes $u'$ and $v$, which results in a sparse matrix $\mathbf{W}_k^{(-)}$. Averaging $\kappa$ times over such adjacency matrices is equivalent to sampling from the negative distribution $p_n(u'|v)$.

Combining Eq.~\eqref{eq:pos} and~\eqref{eq:neg} gives Eq.~\eqref{eq:CL}. 
\comment{
\begin{equation}
    \argmax\limits_{\mY,\text{ s.t. }\mY^\top\!\mY=\mIdent} \trace(\mathbf{Y}^\top\!\Delta\mathbf{W}\mathbf{Y}) \quad\text{ where }\quad \Delta\mathbf{W}\!=\!\mathbf{W}^{(+)}-\frac{\eta'}{\kappa}\sum\limits_{k=1}^\kappa\!\mathbf{W}_k^{(-)}.
\end{equation}
where the constraint $\mY^\top\!\mY=\mIdent$ is designed for eliminating the trivial solution.}

\paragraph{Block-Contrastive Loss.}
Based on Eq.~\eqref{eq:pos1}, we can extend  Eq.~\eqref{eq:block} into two different items:
\begin{equation}
\mathbb{E}_{u \sim p_{d}(u \mid v)} (\mathbf{u}^{\top} \mathbf{v}) =  \mathbf{v}^\top\sum_{u} \frac{\widehat{W}_{uv}}{\sqrt{D_{vv}D_{uu}}}\mathbf{u}= \mathbf{v}^\top\sum_{u} W^{(+)}_{uv}\mathbf{u},
\end{equation}
and
\begin{equation}
\mathbb{E}_{u'\sim p_{n}\left(u'\mid v\right)} \left(\mathbf{u}'^\top\mathbf{v}\right) = \mathbf{v}^\top\sum_{u'} W^{(-)}_{u'v}\mathbf{u'}.
\end{equation}
Thus, we have $\boldsymbol{\mu}^+ = 
\sum_{u} W^{(+)}_{uv}\mathbf{u}$ and $\boldsymbol{\mu}^-=
\sum_{u'} W^{(-)}_{u'v}\mathbf{u'}$ in our case. For brevity, we omit $b$ in the above result, whose role in  Eq.~\eqref{eq:block} is to normalize by the block size \eg, the number of links between $v$ and $u$ (and some $b'$ for $v$ and $u'$, respectively). Based on the above derivations, Eq.~\eqref{eq:block} can be reformulated as:
\begin{equation}
    \!\!\!\!-\mathbb{E}_{u \sim p_{d}(u \mid v)} (\mathbf{u}^{\top} \mathbf{v})+\mathbb{E}_{u'\sim p_{n}\left(u'\mid v\right)} \left(-\mathbf{u}'^\top\mathbf{v}\right)=-\mathbf{v}^\top(\boldsymbol{\mu}^{+}-\boldsymbol{\mu}^{-}) = -\mathbf{v}^\top\sum_{u} (W^{(+)}_{uv} -W^{(-)}_{uv})\mathbf{u}.
\end{equation}
Thus, $\sum_{u,v} (W^{(+)}_{uv} -W^{(-)}_{uv})\mathbf{u}^\top\mathbf{v}$ with the corresponding matrix form $\trace(\mathbf{Y}^\top(W^{(+)} -W^{(-)})\mathbf{Y})$. 

\section{Graph Homophily Predicts that COLES Outperforms SampledNCE with Sigmoid (an Intuitive Illustration)}

Let us define the graph homophily for graph $G^{(+)}$ with the degree-normalized adjacency matrix $\mW^{(+)}$, $n$ nodes and multiclass labels $l_1,\cdots,l_n$ as:
\begin{equation}
\mathcal{H}\big(G^{(+)}\big)=\frac{1}{n} \sum_{i=1}^n \frac{1}{|\mathcal{N}_i|} \sum_{j \in \mathcal{N}_i }\delta(l_i - l_j)=\frac{1}{n} \sum_{i=1}^n \sum_{j=1}^n W^{(+)}_{ij}\delta(l_i - l_j),
\end{equation}
where $\delta(l_i - l_j)$ equals one if $l_i$ equals $l_j$, zero otherwise.

Furthermore, for negative sampling, we use the so-called negative graph $G^{(-)}$, which is a sparse graph with the uniform probability $p'>0$ of connection between each pair of nodes. Thus, in expectation, the homophily of this graph is equal to homophily for the fully-connected graph, and is given by: 
\begin{equation}
\mathcal{H}\big(G^{(-)}\big)=\frac{1}{n} \sum_{i=1}^n \sum_{j=1}^n W^{(-)}_{ij}\delta(l_i - l_j)=\frac{1}{C}\sum_{c=1}^C\rho^{2}_c,
\end{equation}
where $C$ is the number of classes, $\rho_1,\cdots,\rho_C$ are class probabilities \eg, $\rho_1=0.1$ means that class one is given to the 10\% of nodes. 

Looking at Eq. \eqref{eq:ganlike}, we notice that for the  SampledNCE with sigmoid, one can think of $D(x)$ and $1-D(x')$ as a sigmoid for $x p_r$ and a reverse sigmoid for $p_g$, respectively. Therefore, to understand how well two distributions are separated, one can measure:
\begin{equation}
\mathrm{Sep}(v, u, u')=\frac{|D(\vu^\top\vv)-D(\vu'^\top\vv)|}{D(\vu^\top\vv)+D(\vu'^\top\vv)},
\end{equation}
where $\mathrm{Sep}(v, u, u')\rightarrow 1$ if the dot-products of embeddings $\langle\vv,\vu\rangle$ and $\langle\vv,\vu'\rangle$ can be separated from each other linearly, and $Sep(v, u, u')\rightarrow 0$ if they cannot be separated.

To this end, we make a simple assumption. If $\mathcal{H}\big(G^{(+)}\big)\rightarrow \mathcal{H}\big(G^{(-)}\big)$, this means that $p_r$ and  $p_g$ become highly similar, which is good for the underlying JS divergence but it means that is impossible to find embeddings which will separate two distributions (contrastive learning fails in this regime). At the other extreme end, $\mathcal{H}\big(G^{(+)}\big)\gg \mathcal{H}\big(G^{(-)}\big)$, which indicates that we can easily find embeddings that separate  $p_r$ and  $p_g$. However, these embeddings can be disjoint, which is manageable for the underlying surrogate of Wasserstein distance ins COLES but is hard for SampledNCE with sigmoid with the underlying JS divergence.

\begin{figure}[!htbp]
 \centering
\includegraphics[width=8cm]{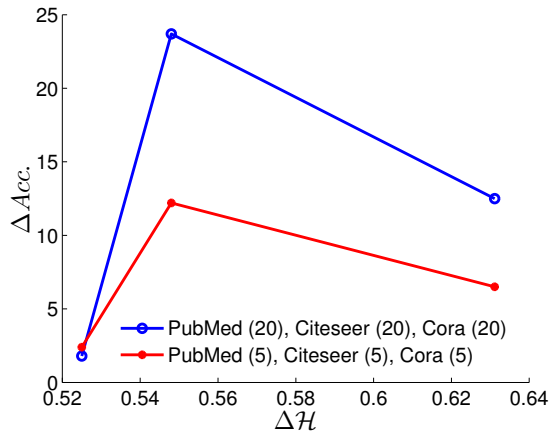}
\caption{$\Delta \mathrm{Acc.}$ between  COLES-GCN and GCN+SampledNCE \wrt $\Delta\mathcal{H}$. We sorted the results by $\Delta\mathcal{H}$. Thus, the first, second and third point on each curve (from left to right) corresponds to Pubmed, Citeseer and Cora, respectively. In parentheses, we indicate the number of labeled training samples per class.}
\label{fig:hom}
\end{figure}

To validate our intuition, Figure \ref{fig:hom} shows $\Delta \mathrm{Acc.}$ between COLES-GCN and GCN+SampledNCE as a function of $\Delta\mathcal{H}=\mathcal{H}\big(G^{(+)}\big)-\mathcal{H}\big(G^{(-)}\big)$. We use the same experimental setting as the one used for results reported in Table \ref{tab:trans}. After sorting results by homophily in the ascending order, we note that the overall trend agrees with our expectations that for small $\Delta \mathcal{H}$, both methods struggle more as it is harder for the contrastive setting to find distinctive embeddings. However, as $\Delta \mathcal{H}$ increases, the overlap between  $p_r$ and  $p_g$ decreases, making it easier to find distinctive embeddings. COLES benefits a lot under this setting, whereas SampledNCE with sigmoid benefits to a lesser degree. 

The above simple illustration/intuition is by no means an exhaustive proof given we evaluated it on only three datasets, and embeddings can exploit often complex neighborhood patterns which the homophily index cannot capture (something appearing as random from the homophily perspective may still enjoy an informative complex pattern). Nonetheless, our observation  supports our claim that COLES works well in the regime where contrastive learning is easily viable, whereas the SampledNCE with sigmoid struggles more by contrast.

\comment{
\section{Relation between Contrastive Loss and Graph Homophily}
In this paper, we introduce GAN to explain the issue of contrastive loss. We define discriminator $D(x)=\sigma(x)$ where $x=\mathbf{u}^{\top}\mathbf{v}$ if $x$ is sampled from the 'real' distribution $p_r = p_{d}(u \mid v)$ and $x=\mathbf{u}'^{\top}\mathbf{v}$ if $x$ is sampled from the 'generator' distribution $p_g = p_{n}\left(u' \mid v\right)$.
Here we try to explain what are the $p_r = p_{d}(u \mid v)$ and $p_g = p_{n}\left(u' \mid v\right)$ exactly because we do not have the 'real' and 'fake' samples like GANs. 
In contrast 'real' and 'fake' samples, if we regard a pair like a sample we could have neighbourhood pairs and non-neighbourhood pairs.
In the contrastive loss, we use neighbourhood pairs and non-neighbourhood pairs to construct positive and negative pairs. And thus our 'real' samples are scores from neighbourhood pairs and 'fake' samples are scores from non-neighbourhood pairs. Based on this, we can understand that the discriminator $D(x)$ plays a role in determining a given pair is neighbour pair or not from the score $x$.
Thus, we need to pay attention to the scores from positive pairs and negative pairs. Please note that a positive pairs do not mean a pair should belong to the same class while a negative pairs do not mean a pair should not belong to the same class. Further, we will explain what kinds of factors will make this situation happen. There are two different factors affecting scores from positive pairs and negative pairs respectively: homophily and class distribution.

\paragraph{Graph Homophily.} The homophily principle~\cite{mcpherson2001birds} in the context of node classification asserts that nodes from the same class tend to form edges. Homophily is also a common assumption in graph clustering~\cite{von2007tutorial} and in many GNNs design~\cite{klicpera2018predict}.  GNNs model
the homophily principle by aggregating node features within graph neighborhoods. %
In \cite{pei2020geom}, researcher propose a index to measure homophily:
\begin{equation}
\mathcal{H}(G)=\frac{1}{|V|} \sum_{v \in V} \frac{ \sum_{u \in N_v }(\mathbf{y}_u = \mathbf{y}_v)}{|N_v|},
\end{equation}
where $\mathbf{y}_v$ is the one-hot label for node $v$. and the $\sum$ will account the number of neighbors of $v\in V$ that have the same label as $v$. We can observe this definition give an index to the boundary between two classes.

It is easy to know when we have a graph $G$ with $\mathcal{H}(G)=1$ the positive pairs will become the real positive pairs because nodes from each neighbourhood pairs will belong to the same class. We also consider the setting $\mathcal{H}(G)=0$, then the positive pairs are worse than the random pairs because even random pairs probably belong to the same class.
If we do propagation with a graph of $\mathcal{H}(G)=1$, intra-distance are close to zeros as increasing propagation times. That means even if the raw features do not have any discriminative power (inseparable in linear or nonlinear classifiers), it is still possible to obtain discriminative embeddings based on the information in the graph itself.

\paragraph{Class Distribution.} For the negative pairs, although we use the name 'negative' they are not always negative and highly dependent on class distribution. For a given node $u$ with $y_u = c$, if the samples with class $c$ account for p\% of the total samples, there is p\% probability that the negative pair with the node $u$ belong to the same class while there is 1-p\% probability that the negative pair with the node $u$ belong to the different class.
We start the story from an example with balanced class distribution as shown in Fig.~\ref{fig:uniform}.
     \begin{figure}
         \centering
         \includegraphics[width=10cm]{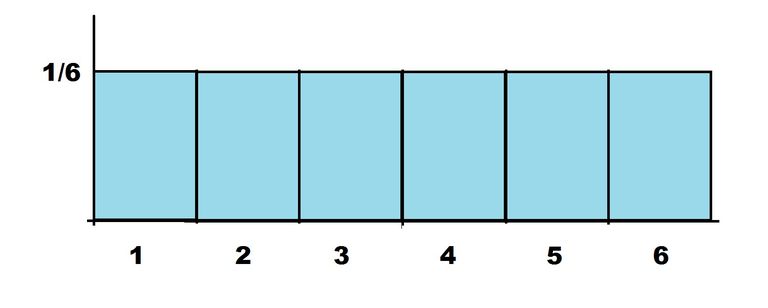}
         \caption{Balanced Class Distribution}
         \label{fig:uniform}
     \end{figure}
Given a nodes, we want to sample an other non-neighbourhood nodes to be a pair. In this case, there is only a 1/6 probability that the nodes in the negative pair belong to the same class while there is 5/6 probability that the nodes in the negative pair belong to the different class. In uniform class distribution, if the number of categories $n\rightarrow\infty$, almost all negative pairs have the right labels.

For unbalance class distribution, we assume the biggest class accounts for $p_{big}$\% while the smallest class accounts for $p_{small}$\%. We could have the upper bound $p_{big}\%\times p_{big}\%$ and lower bound $p_{small}\%\times p_{small}\%$ for the error rate of negative pairs.

\paragraph{Combination of homophily and class distribution} 
We can easily see that there are two extreme cases. One is that $\mathcal{H}(G)=1$ while the number of categories tends to be infinite. In this case, the high homophily will reduce intra-class distances for each class. If the class center of each class is far from the class centers of the rest classes, the fake scores only depend on the inter-class distance while the real scores will be a constant. That means the distributions of real scores and fake scores tend to be disjoint. And Janson-Shannon divergence cannot work well in this situation.

Another case is that $\mathcal{H}(G)=0$ while the number of categories is small and imbalance (e.g. two classes and 90\% of the samples are in the same category). $\mathcal{H}(G)=0$ determines that we can not have any helpful information from the graph to improve the discrimination by reducing intra-distance. At the same time, the imbalance class distribution results in many wrong negative pairs ($90\%\times90\%$ negative pairs are wrong.), which leads to the deterioration of learning process.
}

\section{Reproducibility}
\subsection{Datasets}
In this paper we use six datasets to evaluate our method. 
Cora is a well-known citation network labeled according to the paper topic. Most of approaches report on a small subset of this dataset. The \textbf{Cora} dataset consists of 2708 scientific publications classified into one of seven classes. The citation network consists of 5429 links.
Each publication in the dataset is described by a 0/1-valued word vector indicating the absence/presence of the corresponding word from the dictionary. The dictionary consists of 1433 unique words. \textbf{Cora Full} consists of 19793 scientific publications classified into one of seventy classes. The citation network consists of 65311 links.  The dictionary consists of 1433 unique words. 

The \textbf{CiteSeer} dataset consists of 3312 scientific publications classified into one of six classes. The citation network consists of 4732 links. Each publication in the dataset is described by a 0/1-valued word vector indicating the absence/presence of the corresponding word from the dictionary. The dictionary consists of 3703 unique words.

The \textbf{Pubmed} dataset consists of 19717 scientific publications from PubMed database pertaining to diabetes classified into one of three classes. The citation network consists of 44338 links. Each publication in the dataset is described by a TF/IDF weighted word vector from a dictionary which consists of 500 unique words.

The \textbf{Reddit} dataset is a graph dataset from Reddit posts made in the month of September, 2014. The node label in this case is the community, or “subreddit”, that a post belongs to. The 50 large communities have been sampled to build a post-to-post graph, connecting posts if the same user comments on both. In total, this dataset contains 232,965 posts with an average degree of 492. The first 20 days are used for training and the remaining days for testing (with 30\% used for validation).

The \textbf{Ogbn-arxiv} dataset contains a directed graph, representing the citation network between all Computer Science (CS) arXiv papers indexed by MAG~\cite{wang2020microsoft}. Each node is an arXiv paper and each directed edge indicates that one paper cites another one. Each paper comes with a 128-dimensional feature vector obtained by averaging the embeddings of words in its title and abstract. The embeddings of individual words are computed by running the skip-gram model~\cite{mikolov2013distributed} over the MAG corpus. We also provide the mapping from MAG paper IDs into the raw texts of titles and abstracts here. In addition, all papers are also associated with the year that the corresponding paper was published.

\section{Implementation}
We use PyTorch to implement COLES and its variants.%
The propagation procedure is efficiently implemented with sparse-dense matrix multiplications. The codes of GCN, COLES-GCN, SGC, COLES-SGC, S\textsuperscript{2}GC and COLES-S\textsuperscript{2}GC are also implemented with PyTorch.
The weight matrices of classifier are initialized with Glorot normal initializer.
We employ Adam~\cite{kingma2014adam} to optimize parameters of the proposed methods and adopt early stopping to control the training epochs based on validation loss.
For the experiments on Cora, Citeseer, Pubmed, CoraFull, we use SGD to optimize Eq.~\eqref{eq:CL} because the datasets are small enough. For reddits and Ogbn-arxiv, we use Eq.~\eqref{eq:COLES_linear} to obtain the closed-form solution to accelerate the speed. 
All the experiments in this paper are conducted on a single NVIDIA GeForce RTX 1080 with 8 GB memory. Server operating system is Unbuntu 18.04. As for software versions, we use Python 3.7.3, PyTorch 1.6.0, NumPy 1.18.1, SciPy 1.4.1, CUDA 9.1.

\subsection{Hyperparameters}
We did not put much effort to tune these hyperparameters in practice, as we observe that COLES is not very sensitive to different hyperparameters. SGC and S\textsuperscript{2}GC use the aggregation step $K$, the only hyperparameter for these methods. Thus we use $K=8$ for most benchmarks. Except for Ogbn-arxiv, we use logistic regression as the classifier for all contrastive based methods. Note that we do not tune any parameter for the logistic regression and just use the default setting. In Ogbn-arxiv, the given features are non-linear because they are based on Bag-of-Words with word embeddings. Thus, the MLP classifier is selected for COLES-S\textsuperscript{2}GC. Specifically, we keep the setting of the MLP classifier in the baseline. There are two hidden state layers, and the hidden state size is 256 dimension for each layer. The learning rate for the MLP is 0.005 and the dropout rate is 0.4.

\begin{table}[]
\centering
\caption{The hyperparameters of datasets (node classification).}
\begin{tabular}{l|llllll}
Dataset   &Optimizer & K   & lr      & weight decay & Epoch & hidden size \\
\hline
Cora      &Adam    & 8      & 1e-3    & 5e-4     &20  &512\\
Citeseer  &Adam    & 8      & 1e-4    & 1e-4     &80  &512\\
Pubmed    &Adam   & 8     & 2e-2       & 1e-5    &40  &256 \\
Cora Full &Adam   & 2     & 1e-2    & 0     &30  &512\\
Ogbn-arxiv &SVD & 10	& None  & None      & 500   &126    \\
Reddit    &SVD & 2 & None & None      & None    &600  \\
\end{tabular}
\label{tab:hyper1}
\end{table}

\begin{table}[]
\centering
\caption{The hyperparameters of datasets (node clustering).}
\begin{tabular}{l|llllll}
Dataset   &Optimizer & K   & lr      & weight decay & Epoch & hidden \\
\hline
Cora      &Adam    & 8      & 1e-2    & 5e-4     &1  &512\\
Citeseer  &Adam    & 8      & 1e-4    & 1e-4     &30  &512\\
Pubmed    &Adam   & 8     & 2e-2       & 1e-5    &40  &256
\end{tabular}
\label{tab:hyper2}
\end{table}

\end{document}